%% file: conference_101719.tex
\documentclass[conference]{IEEEtran}
\IEEEoverridecommandlockouts
\usepackage{cite}
\usepackage{amsmath,amssymb,amsfonts}
\usepackage{listings}
\usepackage{algorithm}
\usepackage{algpseudocode}
\usepackage{graphicx}
\usepackage{subfigure}
\usepackage{cite}
\usepackage{comment}
\usepackage{multirow}
\usepackage{tcolorbox}
\usepackage{tabularx}
\usepackage{eqparbox}
\usepackage{adjustbox}
\usepackage{float}
\usepackage{xcolor}
\usepackage{mathabx}
\usepackage{array}
\usepackage{mdwmath}
\usepackage{url}
\usepackage{rotating}
{}
{}
{}
\usepackage{adjustbox}
\usepackage{placeins}
\usepackage{booktabs}
\newcommand{\bftab}{\fontseries{b}\selectfont}
\usepackage{siunitx}
\usepackage{graphicx}
\usepackage{tikz}
\usetikzlibrary{positioning}
\usepackage{smartdiagram}

\definecolor{as_color}{rgb}{0.823, 0.411, 0.117}

\def\BibTeX{{\rm B\kern-.05em{\sc i\kern-.025em b}\kern-.08em
    T\kern-.1667em\lower.7ex\hbox{E}\kern-.125emX}}

\colorlet{punct}{red!60!black}
\definecolor{background}{HTML}{EEEEEE}
\definecolor{delim}{RGB}{20,105,176}
\colorlet{numb}{magenta!60!black}
\lstdefinelanguage{json}{
    basicstyle=\normalfont\ttfamily,
    numbers=left,
    numberstyle=\scriptsize,
    stepnumber=1,
    numbersep=8pt,
    showstringspaces=false,
    breaklines=true,
    frame=lines,
    backgroundcolor=\color{background}
    ,
    literate=
     *{0}{{{\color{numb}0}}}{1}
      {1}{{{\color{numb}1}}}{1}
      {2}{{{\color{numb}2}}}{1}
      {3}{{{\color{numb}3}}}{1}
      {4}{{{\color{numb}4}}}{1}
      {5}{{{\color{numb}5}}}{1}
      {6}{{{\color{numb}6}}}{1}
      {7}{{{\color{numb}7}}}{1}
      {8}{{{\color{numb}8}}}{1}
      {9}{{{\color{numb}9}}}{1}
      {:}{{{\color{punct}{:}}}}{1}
      {,}{{{\color{punct}{,}}}}{1}
      {\{}{{{\color{delim}{\{}}}}{1}
      {\}}{{{\color{delim}{\}}}}}{1}
      {[}{{{\color{delim}{[}}}}{1}
      {]}{{{\color{delim}{]}}}}{1},
}

\begin{document}

\title{Multi-Source Data Fusion for 
Cyberattack Detection in Power Systems\\
}
\author{\IEEEauthorblockN{Abhijeet Sahu\IEEEauthorrefmark{1},
Zeyu Mao\IEEEauthorrefmark{1},
Patrick Wlazlo\IEEEauthorrefmark{1},
Hao Huang\IEEEauthorrefmark{1}, 
Katherine Davis\IEEEauthorrefmark{1},\\
Ana Goulart\IEEEauthorrefmark{1}, and
Saman Zonouz\IEEEauthorrefmark{1}}
}

\maketitle
\thispagestyle{plain}
\pagestyle{plain}
\begin{abstract}
Cyberattacks can cause a severe impact on power systems unless detected early. However, accurate and timely detection in critical infrastructure systems presents challenges, e.g., due to zero-day vulnerability exploitations and the cyber-physical nature of the system coupled with the need for high reliability and resilience of the physical system. Conventional rule-based and anomaly-based intrusion detection system (IDS) tools are insufficient for detecting zero-day cyber intrusions in the industrial control system (ICS) networks. Hence, in this work, we show that fusing information from multiple data sources can help identify cyber-induced incidents and reduce false positives. Specifically, we present how to recognize and address the barriers that can prevent the accurate use of multiple data sources for fusion-based detection. We perform multi-source data fusion for training IDS in a cyber-physical power system testbed where we collect cyber and physical side data from multiple sensors emulating real-world data sources that would be found in a utility and synthesizes these into features for algorithms to detect intrusions. Results are presented using the proposed data fusion application to infer False Data and Command injection-based Man-in- The-Middle (MiTM) attacks. Post collection, the data fusion application uses time-synchronized merge and extracts features followed by pre-processing such as imputation and encoding before training supervised, semi-supervised, and unsupervised learning models to evaluate the performance of the IDS. A major finding is the improvement of detection accuracy by fusion of features from cyber, security, and physical domains. Additionally, we observed the co-training technique performs at par with supervised learning methods when fed with our features.  
\end{abstract}


\input{Content/1_introduction}
\input{Content/2_background}

\input{Content/3_architecture}

\input{Content/4_multisource}
\input{Content/6_fusion_types}
\input{Content/5_data_transformation}
\input{Content/7_ids}
\input{Content/8_results}

\input{Content/10_conclusion}

\section*{Acknowledgment}
This research is supported by the US Department of Energy's (DoE) Cybersecurity for Energy Delivery Systems program under award DE-OE0000895.

\bibliographystyle{IEEEtran}
\bibliography{reference.bib}


\end{document}

%% file: Content/1_introduction.tex
\section{Introduction}
Multi-sensor data fusion is a widely-known research area adopted in many areas including military, medical science, and finance as well as in the energy sector. Recently, automatic driving systems widely use data fusion to fuse images and videos from similar or disparate sensor types~\cite{auto_vehicle}. In power systems, most fusion applications are currently intradomain and consider only physical data. Examples include fault detection~\cite{pure_phy} and intrusion detection using Principal Component Analysis (PCA)~\cite{pure_phy2}. Similarly, for network protection in industrial control systems (ICS), intrusion detection systems (IDS) such as Snort, BRO, Suricatta, etc., are increasingly used. These offer a pure cyber-centric approach that results in high false alarms~\cite{false_alarms} Combining the benefits of visibility of both cyber and physical, cross-domain data fusion has the potential to help methodically and accurately detect mis-operations and measurement tampering in power systems caused by cyber intrusions.

In power system operations, the telemetry used for collecting wide area measurements may have errors due to sensor damage or cyber-induced compromise; if undetected, applications that rely on these data can become unreliable and/or untrustworthy. Sensor verification based on multi-source multi-domain measurement collection and fusion can be performed to solve such problems, and it is a valuable mechanism for detection and detailed forensics of cyber intrusions targeting physical impact.  While offering numerous potential benefits, fusion for attack detection in real-world utility-scale power systems presents challenges that hinder adoption including the creation, storage, processing, and analysis of the associated large datasets. Fortunately, with the proliferation of affordable computing capability for processing high-dimensional data, it is becoming more feasible to deploy fusion techniques for accurately detecting intrusions.  Thus, research is needed to take advantage of these data and computing capabilities and create fusion-based detection techniques that solve this problem.

Cyberattacks often progress in multiple stages, e.g., initiating with a reconaissance phase, executing intrusions and vulnerability exploitations, and culminating in actions targeting the physical system such as manipulating measurements and commands. 
%
The events that comprise these incidents and provide forensics about what occurred are not reflected using only coarse cyber-side features.
Additionally, the system dynamics in both cyber and physical space vary considerably; this causes challenges in merging data. For example, an intruder may take months in the reconaissance phase, but during this period, none of the physical side features reflect any abnormality. Similarly, later when an intruder is injecting false commands or tampering measurements, most of the cyber side features do not reflect any abnormality, assuming the adversary is stealthy.

Sensor time resolution varies across domains and within domains, which causes challenges when merging the data. The resolution of physical measurements depends on polling rates as well as the specifications of the device. For example, phasor measurement units (PMUs) provide GPS synchronized data at subsecond data rates, supervisory data acquisition and control (SCADA) systems provide data on the seconds to minutes time frame, and smart meters deployed residentially may have hourly resolution~\cite{pmustandard}. Relays monitoring system transients have resolution on the order of milliseconds. Similarly, network logs and IDS such as Snort have resolution of milliseconds. Data fusion solutions for cyber-physical power systems must be able to effectively handle the range of time scales.

The use of machine learning (ML) and deep learning (DL) for intrusion detection faces the problem that the trained model's effectiveness depends on the data collected; it is a challenge to obtain a realistic baseline and to use realistic data to validate the solution for a real-time cyber-physical system. Detection is affected by the choice of data processing techniques applied (e.g., balancing, scaling, encoding). The impact of such factors on detection accuracy must therefore be quantified before the techniques can be trusted for use in securing critical infrastructure.

The hypothesis of this work is that the use of fused data from cyber and physical domains can enable better attack detection performance than either domain separately, if the challenges above are addressed. Hence, we present a heterogeneous-source platform that fuses data and detects cyber intrusions. First, we provide interfaces for collecting data sources from cyber and physical side emulators. Then, we use these interfaces to collect real-time data from cyber, physical, and security domains; finally, we fuse the datasets and detect cyber intrusions. We aggregate and merge real-time sensor data from multiple sources including Elasticsearch\cite{elasticsearch}, TShark\cite{tshark}, raw packet captures with DNP3 traffic, and Snort logs \cite{snort_cookbook} that are created during emulation of Man-in-the-Middle attacks on a synthetic electric grid, modeled in the Resilient Energy Systems Laboratory (RESLab) testbed \cite{Sahu2020}.
Fig. \ref{fig:flowchart} gives an overview of the multi-source data fusion presented.  The major contributions of this paper are as follows: 

\begin{figure}
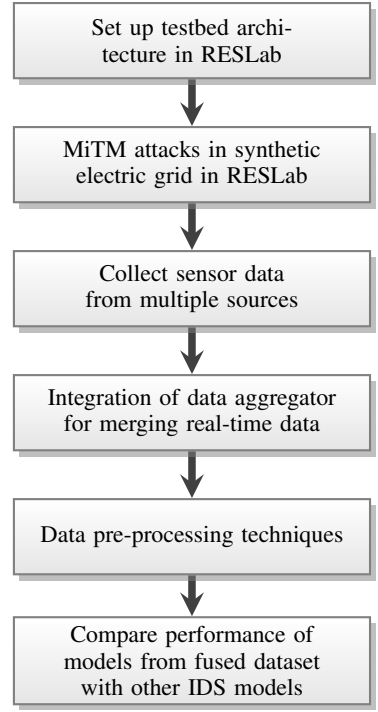

    
\smartdiagramset{text width=4.5cm,
}
\begin{center}
\smartdiagramset{module shape=rectangle,
uniform arrow color=true,
arrow color=gray!50!black,
back arrow disabled=true,
}
\smartdiagramset{
uniform color list=white!90!black for 6 items}
\smartdiagram[flow diagram]{Set up testbed architecture in RESLab,MiTM attacks in synthetic electric grid in RESLab,{Collect sensor data from multiple sources}, Integration of data aggregator for merging real-time data,Data pre-processing techniques,
Compare performance of models from fused dataset with other IDS models}
\end{center}
\caption{Multi-source data fusion steps}\label{fig:flowchart}
\end{figure}

\begin{enumerate}
    \item To present the aggregation and merging of real-time sensor data from multiple sources for cyberattack detection in a cyber-physical testbed emulation of a synthetic electric grid.
    \item To quantify the value of different data pre-processing techniques such as balancing, normalization, encoding, imputation, feature reduction, and correlation before training the machine learning models.
    \item To demonstrate the improved detection capability of models built from fused dataset performance by comparing with pure cyber and physical feature based intrusion detection models. 
    \item To evaluate the performance of the supervised, unsupervised and semi-supervised learning based intrusion detection for use cases explored in the MiTM attacks.
\end{enumerate}

The paper proceeds as follows. Section~\ref{background} provides background on data fusion techniques incorporated in areas such as military, healthcare, software firms, security, and cyber-physical systems. In Section~\ref{architecture}, we discuss the RESLab architecture, the attack types considered, and the data fusion procedure. The details on the data sources, the data fusion types, and the dataset transformations used in this work are presented in Sections~\ref{dataset}, \ref{data_transformation}, and \ref{fusion_types} respectively.
Finally, intrusion detection based on unsupervised, supervised, and semi-supervised learning methods is presented in Section~\ref{ids}.
Experiments are performed for four use cases, and results are analyzed in Section~\ref{results} and finally concluding the paper in Section~\ref{conclusion}.

%% file: Content/2_background.tex
\section{Data Fusion Background}\label{background}
\subsection{Multi-Sensor Data Fusion}
The goal of multi-sensor data fusion is to make better inferences than those that could be accrued from a single source or sensor.  According to \textit{Mathematical Techniques in Multisensor Data Fusion}~\cite{hall_book}, multi-sensor data fusion is defined as \textit{``a technique concerned with the problem of how to combine data from multiple (and possibly diverse) sensors in order to make inferences about a physical event, activity, or situation.''} A data fusion process is modeled in three ways: a) functional, b) architectural, and c) mathematical~\cite{hall_book}. A functional model illustrates the primary functions, relevant databases, and inter-connectivity to perform fusion. It involves primarily filtering, database creation, and pre-processing such as scaling and encoding, etc. An architectural model specifies hardware and software components, associated data flows, and external interfaces~\cite{extra_fusion_ref}. For example, it models the location of the fusion tool in a testbed. The fusion architecture can be three types: centralized, autonomous, and hybrid~\cite{hall_book}. In centralized architectures, either raw or derived data from multiple sensors are fused before they are fed into a classifier or state estimator.  In autonomous architectures, the features extracted are fed to the classifiers or estimators for decision making before they are fused. The fusion techniques used in the second case involve techniques including Bayesian~\cite{intro1} and Dempster Shafer inference~\cite{zadeh}, because these fusion algorithms are fed with the probability distributions computed from the classifiers or the estimators. The hybrid type mixes both centralized and autonomous architectures. The mathematical model describes the algorithms and logical processes.

\begin{figure*} [!htb]
    \centering
    \vspace{-0.2cm}
    \includegraphics[width=0.8\linewidth]{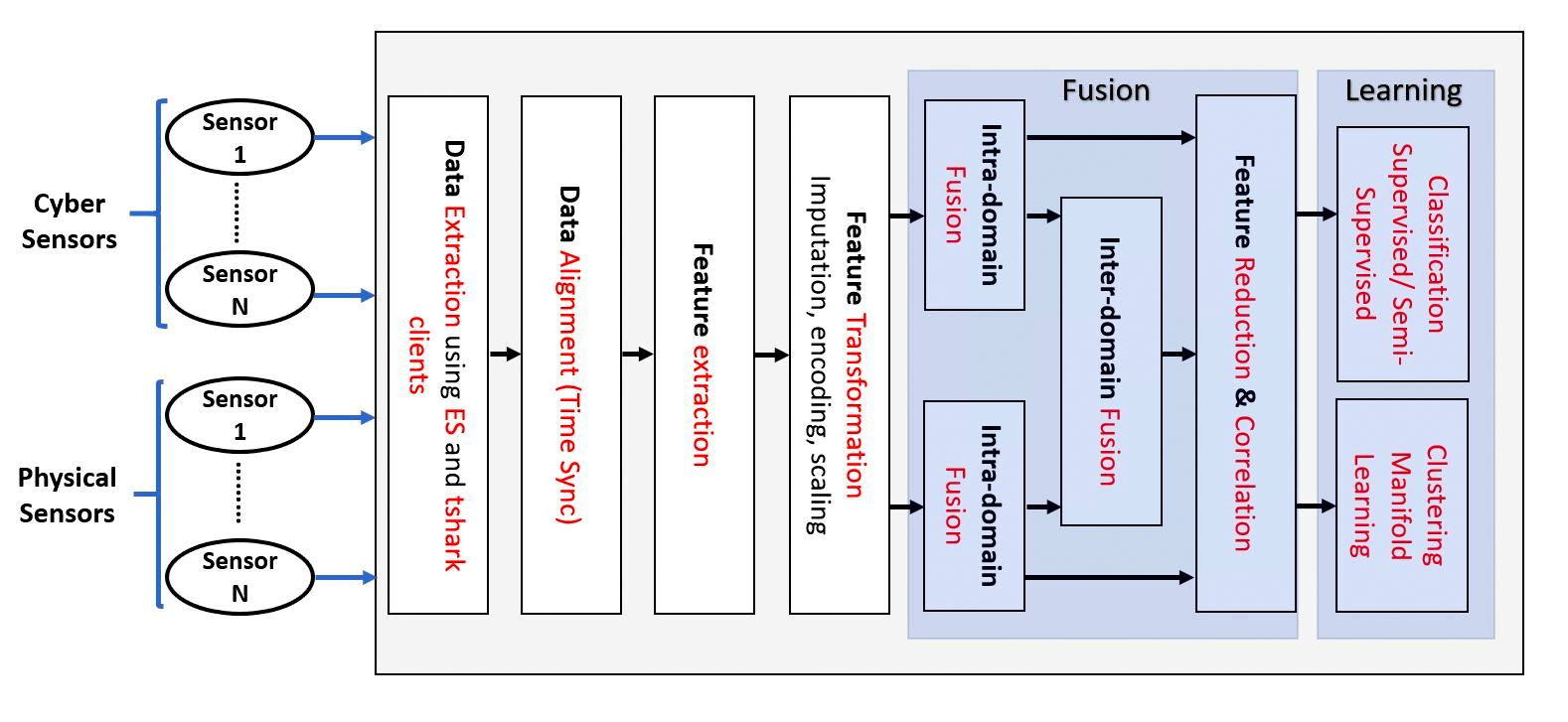}
    \vspace{-0.35cm}
    \caption{Centralized fusion architecture. In the autonomous architecture the Fusion and Learning blocks will be interchanged with an addition of another Learning block post fusion.  
    }
    \label{fig:architecture_types}
\end{figure*}

A holistic data fusion method must consist of all three: functional, architectural, and mathematical models.  The functional model defines the objective of the fusion. Since the goal of this work is to detect intrusions, we must determine which data are due to cyber compromise. Functional goals may also include estimating the position of the intruder in the system or estimating the state of an electric grid, where the preprocessing techniques to use vary based on the goal.  The architecture model defines the sequence of operations. Our fusion follows the centralized architecture. Finally, the mathematical model defines how these features are processed and merged. Section \ref{dataset} details our fusion models.

\subsection{Multi-Sensor Fusion Applications}
Recently, work on multi-sensor fusion has been adopted in the areas of computer vision, automatic vehicle communication, and it is entering into the areas of power systems. The authors in ~\cite{ref3} review multi-sensor data fusion technology, including benefits and challenges of different methods. The challenges are related to data imperfection, outliers, modality, correlation, dimensionality, operational timing, inconsistencies, etc. For example, different time resolutions of sensors result in under-sampling or over-sampling data in some sensors. The response time of certain sensors also vary depending on the sensor age and type.  Data received from multiple sensors must be transformed to a common spatial and temporal reference frame~\cite{hall_book}. Imperfection is dealt using fuzzy set theory, rough set theory, Dempster Shafer theory, etc. 

Multi-sensor data fusion is used in military applications for automated target recognition, battle-field surveillance, and guidance and control of autonomous vehicles~\cite{hall}. Further, the idea has been expanded to non-defense areas such as medical diagnosis, smart buildings, and automatic vehicular communications ~\cite{multi2}. Authors in~\cite{ref4} explore techniques in multi-sensor satellite image fusion to obtain better inferences regarding weather and pollution. Data fusion has also been proposed to accurately detect energy theft from multiple sensors in advanced metering infrastructure (AMI) in power distribution systems~\cite{multi}.

Data fusion is expanded in ~\cite{ref1} from cyber-physical systems (CPSs) to cyber-physical-social systems (CPSSs) with the use of tensors. Algorithms proposed for mining heterogeneous information networks cannot be directly applied to cross-domain data fusion problems; fusion of the knowledge extracted from each dataset gives better results ~\cite{cross_fusion}.

\subsection{Data Fusion in Power Systems}
The data from diverse domains play a major role in power system operation and control. Weather data is vital for forecasting, e.g., for solar, wind, and load, to schedule generation. Data in cyberspace include data that provide for automation in power system ICS and play a crucial role in wide area control and operation in the electric grid. However, to proceed with multi-domain fusion, the following question must first be answered: To what measurable quantities do \textit{cyber data} and \textit{physical data} refer? 

A simple example of \textit{cyber data} in ICS is a spool log of a network printer in the control network.  It is crucial to question, could we have prevented the attack on the centrifuge in the Natanz Uranium Enrichment plant, if we had a logger to record the events of a machine with shared printer, so as to prevent the exploitation of remote code execution on this machine? The answer is \textit{no}, because there were many other vulnerabilities such as WinCC DB exploit, network share, and server service vulnerability, in parallel to print server vulnerability that compromised the Web Navigation Server which was connected to the Engineering Station that configured the S7-315 PLCs which over-speeded the centrifuge~\cite{stuxnet}. Hence, the deployment of cyber telemetry in every computing node in an ICS network is a solution which seems attractive but results in numerous false alarms. Then, the question arises, can we reduce such alerts by amalgamating such data with data from physical sensors? 

Data fusion proposed in the areas of power systems are mainly intra-domain. Existing works do not consider fusion of cyber and physical attributes for intrusion detection together. A probabilistic graphic model based power systems data fusion is proposed in~\cite{pure_phy3}, where the state variables are estimated based on the measurements from heterogeneous sources by belief propagation using factor graphs. These probabilistic models require the knowledge of the priors of the state variables and also assume the measurements to be trustworthy. Hence, such solutions cannot detect cyber induced stealth false data injection attacks. Several works on false data injection detection are based on machine learning~\cite{ml_power_1,ml_power_2,ml_power_3,ml_power_4} and deep learning~\cite{dl_power_1,dl_power_2,dl_power_3,arnav_rnn_lstm,a3d,paved} techniques. The authors in~\cite{pure_phy4} address stealthy attacks using multi-dimensional data fusion by collecting information from power consumption of physical devices, control operation and system states feed to the cascade detection algorithm to identify stealthy attack using Long Short Term Memory (LSTM).   Machine learning techniques including clustering are used in power system security for grouping similar operating states (emergency, alert, normal, etc.) to automatically identify the subset of attributes relevant for prediction of the security class. A decision tree based transient stability assessment of the Hydro-Quebec system is presented in~\cite{ml_power_sys}. Techniques of fusion for fault detection~\cite{pure_phy} and real-time intrusion detection using Principal Component Analysis (PCA)  (PCA)~\cite{pure_phy2} are specific to the physical domain. Design of such models require data fusion and must consider impending system instabilities that can be caused by cyber intrusions.

Cymbiote~\cite{cymbiote} multi-source sensor fusion platform is one of the work, equivalent to ours, that have leveraged fusion from multiple cyber and physical streams and trained with only supervised learning based IDS. Moreover, their work doesnt clearly describes the features extracted from different sources.

\subsection{Multi-Domain Fusion Techniques}

Techniques such as co-training, multiple kernel learning, and subspace learning are used for data fusion problems. Co-training based algorithms~\cite{co_training} maximizes the mutual agreement between two distinct views of the data. This technique is used in fault detection and classification in transmission and distribution systems~\cite{cotrain_1} as well as in network traffic classification~\cite{cotrain_2}.  
To improve learning accuracy, Multiple kernel learning algorithms~\cite{mkl} are also considered, which utilize kernels that implicitly represent different views and combines them linearly or non-
linearly . Subspace learning algorithms~\cite{subspace_learning} aim to obtain a latent subspace shared
by multiple views, assuming that the input views are generated from this latent subspace.
DISMUTE~\cite{dismute} performs feature selection for multi-view cross-domain learning. Multi-view discriminant transfer (MDT)~\cite{mv_transfer_learning} learns discriminant weight vectors for each view to minimize the domain
discrepancy and the view disagreement simultaneously. These techniques can be used for cross-domain data fusion. 

Coupled matrix factorization and manifold alignment methods are used for similarity based data fusion~\cite{cross_fusion}. These methods can be implemented intra-domain with multiple data sources. Manifold alignment is another technique that generate projections between disparate data sources, but assumes the generating process shares a common manifold. 
Since the primary goal in this work is to fuse datasets from inter-domain such methods may not be effective enough. Still we have explored manifold learning for the purpose of feature reduction to train the supervised learning based classifier.

To the best of our knowledge, co-training has not yet been implemented in an intrusion detection system that uses inter-domain fusion. Hence, in this work, we perform co-training in inter-domain fused datasets by splitting the dataset into cyber and physical views.  


\subsection{Data Creation, Storage, and Retrieval}

The storage and retrieval of multi-sensor data play a major role in fusion and learning. A relational database management system (DBMS) is predominantly used in traditional EMS applications. For example, B.C. Hydro proposes a data exchange interface in a legacy EMS and populates a relational database with the schematic of the Common Information Model (CIM) defined in IEC 61970~\cite{rdbms}. With the proliferation of multiple protocols and data from diverse sources, it is difficult to construct the Entity Relationship model of a relational database management system (RDBMS), since the schema cannot be fixed. Since NoSQL stores unstructured or semi-structured data, usually in the key-value pairs or JSON  documents,  NoSQL is  highly  encouraged  to make use of database such as Elasticsearch~\cite{elasticsearch}, MongoDB~\cite{mongodb}, Cassandra~\cite{cassandra}, etc., for multi-sensor fusion with heterogeneous sources.

Creation of multi-domain datasets to advance the research is a challenging task, since it requires development of a cyber-physical testbed that processes real-time traffic from different simulators, emulators, hardware, and software. Currently, few datasets are publicly available that provide features from diverse domains and sources. Most of the datasets are simulator-specific, which restricts the domain to either pure physical or cyber. The widely-known KDD \cite{kdd} and CIDDS\cite{cidds} datasets used in developing ML-based IDS for bad traffic detection and attack classification are centric to features in the cyber domain~\cite{ieee_cqr}. Tools such as MATPOWER \cite{matpower} and pandapower \cite{pandapower} provide datasets for physical-side bad data detection. Datasets that include measurements related to electric transmission systems including normal, disturbance, control, and cyberattack behaviors are presented in~\cite{dataset0,dataset1,dataset2,dataset3}. The datasets contain phasor measurement unit (PMU) measurements, data logs from Snort, and also data from a gas pipeline and water storage tank plant. The features in these datasets lack fine-grained details in the cyber, relay, and control spaces, as all the features are binary in nature. A cyber-physical dataset is presented in~\cite{dataset4} for a subsystem consisting of liquid containers for fuel or water, with its automated control and data acquisition infrastructure showing 15 real-world scenarios; while it presents a useful way of framing the data fusion problem and approaches for cyber-physical systems (CPS), it is not power system specific.     

A problem in training machine learning (ML) or deep learning (DL) models for intrusion detection through classification, clustering, and fine-tuning hyper-parameters is that its effectiveness depends on the data collected. That is, a practical challenge is to obtain a baseline which needs to come from realistic data. Emulation is preferred to simulation for CPS networks since a simulator demonstrates a network’s behavior while an emulator functionally replicates its behavior and produces real data. Using real data is important to validate that ML or DL solutions address the actual challenges faced in the data from a real-time cyber-physical system. 

The performance of ML and DL models is impacted by the choice of data processing techniques applied to the inputs such as balancing, scaling, or encoding before training the models. The effect of these preprocessing techniques needs to be quantified on the outputs of such ML models before they can be trusted for use in industry.

%% file: Content/3_architecture.tex
\section{Data Fusion Architecture}\label{architecture}
Before discussing the data fusion procedures, it is essential to understand the architecture of the RESLab testbed that is producing the data during emulation of the system under study. 

\subsection{Testbed Architecture}

The RESLab testbed consists of a network emulator, a power system emulator, an OpenDNP3 master and a RTAC based master, an intrusion detection system, and data storage, fusion and visualization software. A brief overview of each component is given below. The detailed explanation of RESLab including its architecture and use cases is provided in ~\cite{Sahu2020}.
\begin{itemize}

  \item \emph{Network Emulator} - Common Open Research Emulator (CORE) is used to emulate the communication network that consists of routers, linux servers, switches, firewalls, IDSes and bridges with other components emulated with other virtual machines (VMs) in vSphere environment. 
  
  \item \emph{Power Emulator} - Power World Dynamic Studio (PWDS) is a real-time simulation engine for operating the simulated power system case in real-time as a DS server~\cite{powerworld}. It is used to simulate the substations in the Texas 2000 case as DNP3 outstations. \cite{synthetic_comm}.

  \item \emph{DNP3 Master} - DNP3 Masters are incorporated using an open DNP3 based application (both GUI and console based) and a SEL-3530 Real-Time Automation Controller (RTAC) that polls measurements and operates outstations, sending its traffic through CORE to the emulated outstations in PowerWorld DS.  

  \item \emph{Intrusion Detection System} - Snort is used in the testbed as the rule-based, open-source intrusion detection system (IDS). It is configured to generate alerts for Denial of Service (DoS), MiTM, and ARP cache poisoning based attacks. Currently Snort is running as a network IDS in the router in the substation network.
  
  \item \emph{Storage and Visualization} - The Elasticsearch, Logstash, and Kibana (ELK) stack is used to probe and store all virtual and physical network interface traffic. In addition to storing all Snort alerts generated during each use case, this data is able to be queried using Lucene queries to perform in depth visualization and cyber data correlation.
  
  \item \emph{Data Fusion } - A different VM is dedicated to operate the fusion engine that collects network logs and Snort alerts from ELK stack using an Elasticsearch client and raw packet captures from CORE using pyshark. This engine constructs cyber and physical features and merges them using the time stamps from different sources to ensure correct alignment of information. Further it pre-processes them using imputation, scaling and encoding before training them for intrusion detection using supervised, unsupervised and semi-supervised learning techniques.  This VM is equipped with resources to utilize ML and DL based library such as scikit, Tensorflow and Keras to train the engine for classification, clustering and inference problem. 
  
\end{itemize}

\begin{figure} [h]
    \centering
    \vspace{-0.2cm}
    \includegraphics[width=1.0\linewidth]{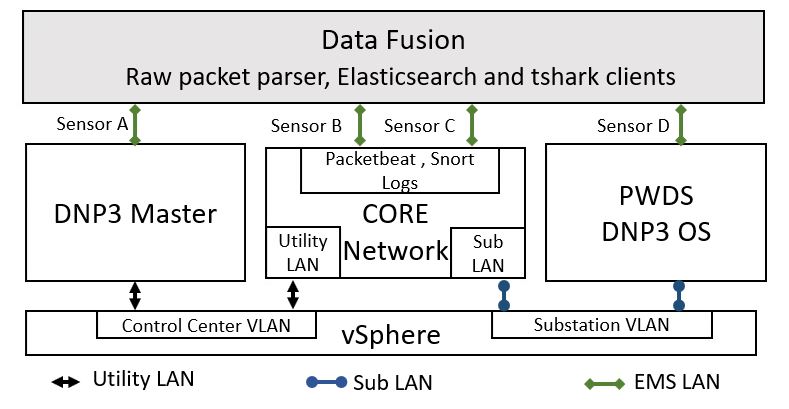}
    \vspace{-0.35cm}
    \caption{Testbed architecture with data fusion}
    \label{fig:architecture}
\end{figure}

There can be considered three broad kinds of IDS for ICS: protocol analysis based IDS, traffic mining based IDS, control process based IDS~\cite{pure_cy_pure_phy}. The fusion engine in RESLab combines all these types. It performs protocol specific feature extraction from data link, network, transport layers alongwith DNP3 layer, control and measurement specific information through DNP3 payload and headers, traffic mining by extracting network logs from multiple sources.

\subsection{Attack Experiments}

Now that we have discussed the architecture of the testbed, we delve further into how this testbed is utilized to demonstrate a few cyber attacks targeting the grid operation. The threat model we consider is based on emulating multi-stage attacks in a large-scale power system communication network. In the initial stage, the adversary gains access to the substation LAN through Secure Shell (SSH) access, further performing DoS and ARP cache poisoning based MiTM attack to cause False Data injections (FDI) and False Command injections (FCI).

In Man-in-the-Middle attacks, usually the adversary secretly observes the communication between sender and receiver and sometimes manipulates the traffic between both ends. There are different ways to perform MitM such as IP spoofing, ARP spoofing, DNS spoofing, HTTPS spoofing, SSL hijacking, stealing browser cookies, etc. In this current work, we focus on MitM using ARP spoofing. ARP spoofing or poisoning is a type of attack in which an adversary sends false ARP messages over a local network (LAN). This results in the linking of an adversary's MAC address with the IP address of a legitimate machine on the network (in our case, the outstation VM). This attack enables the adversary to receive packets from the master as an impersonator for the outstation and modify commands and forward them to the outstation. In this way, the adversary can cause contingencies such as misoperation of the breakers. The attack is not only to modify but also to sniff the current state of the system since it can receive the outstation response to the master. 

The MiTM attacks are performed considering the four use cases targeting different part of the Texas synthetic grid following different strategies presented in detail in ~\cite{Sahu2020}. The use cases are combinations of FDI and FCI attacks performed with different polling rates from the DNP3 Master and the number of master application considered. In our previous work, we demonstrated Snort IDS based detection which resulted in many false positives. In this work, we employ fusion techniques, along with machine learning techniques, to enhance the accuracy of detection by evaluating them using F1-scores, Recall, and Precision values.


\subsection{Data Fusion Procedure}
The steps followed in the data fusion engine, from extracting the features from different sources, with their merge of pyshark, snort, packetbeat, raw packet capture to form cyber table, and the final fusion of cyber and physical table, with the steps of imputation, encoding and visualization is presented in Alg.~\ref{alg:data_fusion}. The details of the sensor sources and the data processing will be discussed in details in the next sections.

\begin{algorithm}[h]
\begin{small}
  \caption{Data Fusion Procedure}\label{alg:data_fusion}
	\begin{algorithmic}
	\State{1. Load json from raw pcaps.}
	\State{2. Extract cyber features: network, transport, datalink layer information and store as raw cyber data.}
	\State{3. Extract features using pyshark.}
	\State{4. Merge pyshark to the raw cyber data.}
	\State{5. Extract snort alert.}
	\State{6. Merge snort to the raw cyber data.}
	\State{7. Extract features from packetbeat index in elasticsearch.}
	\State{8. Merge packetbeat features to raw cyber data.}
	\State{9. Extract DNP3 features (DNP3 points and headers) from raw packet capture.}
	\State{10. Fuse cyber data with physical data.}
	\State{11. Imputate missing values.}
	\State{12. Encode categorical features.}
	\State{13. Visualize the merged table.}
  \end{algorithmic}
  \end{small}
\end{algorithm}

\subsection{Fusion Challenges}
The most challenging task in fusion is to
perform merge operations, because of the different time stamps generated at
different sensors. An event will trigger the time stamped measurements at the sensors. Hence, each
sensor's location impacts the time at which the event is recorded. Domain knowledge
has been used to write the algorithm to merge different sources meticulously. For
example, Elasticsearch’s Packetbeat index stores each record that reflects the
traffic between a given small time interval. Each record has an event start and end
time. While merging Elasticsearch features, such as flow count attribute, we have to
compare the raw packet timestamp and event start and end time of Elasticsearch to
calculate the flow counts.
Moreover, the number of records in the power system side will be less than the
cyber side, as events in power system side are triggered based on the polling
frequency as well as on the time at which an operator performs a control operation.
Hence we remove missing data for the records that do not have any physical side
traffic.

%% file: Content/4_multisource.tex
\section{Multi Sensor Data}\label{dataset}
A sensor's data is the output or readings of a device that detects and responds to changes in the physical environment. Every sensor has a unique purpose that helps create crucial features that can assist in intrusion detection. In RESLab, the cyber sensors are deployed as Wireshark instances at different locations in the network for raw packet capture. Additionally, monitoring tool such as Packetbeat are integrated for extracting network flow-based information. For security sensors, Snort IDS logs and alerts are considered.  Since the physical system is emulated with PWDS acting as a collection of DNP3 outstations, the real-time readings provided by physical sensors are extracted from the observed measurements at the DNP3 master, from the application layer of the raw packet captured at the DNP3 master. The extractions of these multiple sensors are explained in detail:

\subsection{Raw pcaps from json}
The packet captures from Wireshark are packet dissected and saved in the json format, which are loaded using the panda data frame. Further, from the json, around 12 features from physical, datalink, network, and transport layer of OSI stack are extracted, as shown in Table~\ref{table:fusion_features}. The features primarily consist of the source and destination IP and MAC addresses, along with the port numbers, flags, and lengths in these layers. 

\subsection{Elasticsearch}
Real-time traffic collection is performed from network interfaces in CORE, using the Packetbeat plugin in the ELK stack. The Packetbeat plugin helps us extract the flow-based information such as \textit{Flow Count}, \textit{Flow Count Final}, \textit{Packets} shown in Table~\ref{table:fusion_features}.
Elasticsearch queries are based on Lucene, the search library from Apache. Kibana is used to visualize the graphs and real time data visualization for packetbeat index. An example query is shown below:
\begin{lstlisting}[language=json,numbers=none]
"query": {
    "bool": {
      "must": [
          { "range": {
            "event.end": {
            "gte": "2020-01-22T00:00:00.000Z",
            "lte": "2020-01-26T00:00:00.000Z"}}
          },
          {"range": {
           "event.duration": {
            "gte": 0,
            "lte": 3000000}}
          },
          {"bool": 
          {"should": [
          {"match": {
            "destination.port": "20000"}},
          {"match": {
            "source.port": "20000"}}
              ]
          }
          },
          {"match":
            {"flow.final": "true"}
          }    
            ]}}
\end{lstlisting}

The above query returns the records with event start time $2020-01-22T00:00:00.000Z$ and end time $2020-01-26T00:00:00.000Z$, and the event duration is within $0-300000$ ms, and the source or destination port is $20000$ (port number associated with DNP3), and the flow is a $final$ flow. Keyword $must$ designates an $AND$ operation, $should$ is an $OR$ operation, and $match$ is an $equals\: to$ operation.
A logstash index is also created in Elasticsearch to store the logs of Snort alerts, which is also extracted along with the packetbeat index.

There are two operations on the response from Elasticsearch: a) $Extraction$ of essential features b) $Merge$ of features to the existing cyber features data frame $cb\_table$ from raw packet captures. 
Each record in the packetbeat index is stored in the form of an event with start and end times. In the extraction phase, we extract the $source.packets$, $flow.id$, $flow.final$, $event.end$, $event.start$, $flow.duration$ features and store them in a new data frame $pb\_table$.
The merge operation of $pb\_table$ into the existing cyber features is non-trivial due to different timestamps in existing features and features from packetbeat. We compute the features in Table~\ref{table:fusion_features} $flow.count$, $flow.final\_count$, and $packets$ using the features of $event.end (end)$, $event.start (start)$ in $pb\_table$ and $Time$ in the $cb\_table$ based on the logical OR of three conditions:
\begin{enumerate}
    \item $Condition\:1:$ add counters if the event start is within the range of current and next records in the $cyber\_table$
    \begin{equation}
        cb\_table[i][t] \leq start \land cb\_table[i+1][t] \geq start
    \end{equation}
    \item $Condition\:2:$ if the event end is within the range of current and next records in the $cyber\_table$.
    \begin{equation}
        cb\_table[i][t] \leq end \land cb\_table[i+1][t] \geq end
    \end{equation}
    \item $Condition\:3:$ if the event start is less than the current record and event end is greater than the next record in the $cyber\_table$.
    \begin{equation}
        cb\_table[i][t] \geq start \land cb\_table[i+1][t] \leq end
    \end{equation}
\end{enumerate}



The $\lor$ and $\land$ are the logical $or$ and $and$ operators respectively.
In this manner, we merge the three features from $pb\_table$ to the $cb\_table$.

\subsection{Pyshark}
Pyshark is a Python wrapper for tshark, allowing python packet parsing using Wireshark dissectors.
Using $Pyshark$ features such as $Retransmissions$ and $Round Trip Time (RTT)$ are obtained. The RTT is the time duration for a signal or message to be sent plus the time it takes for the acknowledgment of that signal to be received. It has been observed that if congestion is created in any location in between the source and destination such as router or switch, the RTT increases. It also increases due to DoS attacks on the servers or any intermediery nodes in the path between source and destination. The $TCP$ based packet follows different retransmission policies based on the TCP congestion control flavour. Hence, the number of $retransmission$ packets observed within a given time frame is an indicator of loss of communication or increased delay. Usually, a sender retransmits a request if it did not receive an acknowledgment after some multiples of a $RTT$, whose multiplicity is dependent on the TCP flavour. The $retransmission$ and $RTT$ features are selected, as features are correlated and directly related to attacks targeting availability and integrity.



\subsection{Snort}
The router inside the CORE emulator runs the Snort daemon based on the specific rules, pre-processors, and decoders enabled in the configuration file to create logs. 
Snort operates in three modes: packet sniffer, packet logger, and intrusion detection system (IDS) modes. We run the snort in the IDS mode.
The alerts generated at the router in the substation network is continuously probed during the simulation. The alerts are recorded in the form of the $unified2$ format as well as pushed to the Logstash index created in Elasticsearch.
Unified2 works in three modes, packet logging, alert logging, and true unified logging. We run Snort in alert logging mode to capture the alerts, timestamped with alert time. Further, the $idstools$ python package is utilized to extract these $unified2$ formatted logs. The Snort configuration determines which rules and preprocessor are enabled. The features extracted are the $alert$,$alert\_type$, and $timestamp$. The merge into the $cb\_table$ is performed based on the $timestamp$ of each Snort record. The record is inserted based on the condition:
\begin{equation}
    cb\_table[i][t] \geq timestamp \leq cb\_table[i+1][t]
\end{equation}


\subsection{Physical Features from DNP3}

    The Distributed Network Protocol version 3 (DNP3) is widely used in SCADA systems for monitoring and control. This protocol has been upgraded to use TCP/IP in its transport and network layer. It is based on the master/outstation architecture, where field devices are at outstations and the monitoring and control is done by the master. DNP3 has its own three layers: a) Data Link Layer, to ensure reliability of physical link by detecting and correcting errors and duplicate frames, b) Transport Layer, to support fragmentation and reassembly of large application payload, and c) Application Layer, to interface with the DNP3 user software that monitors and controls the field devices. Every outstation consists of a collection of measurements such as breaker status, real power output, etc., which are associated with a DNP3 point and classified under one of the five groups: binary inputs (BI), binary outputs (BO), analog inputs (AI), analog outputs (AO), and counter input. The physical features consist of the information carried in the headers in the three layers of DNP3, along with the values carried by the DNP3 points in the application layer payload. Every DNP3 payload's purpose is indicated by a header in the application layer called function code (FC). In our simulations, we extract the features with FCs: 1(READ), 5(DIRECT OPERATE), 20 (ENABLE spontaneous message), 21(Disable spontaneous message), and 129 (DNP3 RESPONSE). The details of the features are in Table~\ref{table:fusion_features}.   


\begin{table}[]
\caption{Description of the features used in data fusion.}
\begin{tabular}{|p{0.8cm}|m{6.5cm}|m{0.4cm}|}
\cline{1-3}
 Features & Description & Def   \\ \cline{1-3}
 
Frame Len  &  Length of the frame after network, transport and application header and payload are added and fragmented based on the channel type. For ethernet, the frame length can be max. 1518 bytes, which varies for wireless channels. & 0\\ \cline{1-3}

Frame Prot. & Determines the list of protocols in the layers above link layer encapsulated in the frame. & Nan \\\cline{1-3}

Eth Src &   Unique source MAC address. Crucial for detection in ARP spoof attacks. & 0\\\cline{1-3}

Eth Dst &  Unique destination MAC address. Crucial for detection in ARP spoof attacks. & 0\\\cline{1-3}

IP Src &  Unique source IP address. & 0\\ \cline{1-3}

IP Dst &  Unique destination IP address. & 0\\\cline{1-3}

IP Len &  Stores the length of the header and payload in a IP-based packet. This correlates well with the DNP3 payload size. & 0\\ \cline{1-3}

IP Flags & Indicator of fragmentation caused due to link or router congestion in the intermediary nodes. & 0x00\\ \cline{1-3}

Src Port &  Indicates the port number used by the source application using TCP in transport layer. Ex: if the source is the DNP3 outstation, default port is 20000. & 0\\ \cline{1-3}

Dest Port &   Indicates the port number used by the destination application using TCP in transport layer. & 0\\\cline{1-3}

TCP Len &  Stores the length of the header and payload in a TCP-based segment. This correlates well with the DNP3 payload size. & 0\\ \cline{1-3}

TCP Flags &   Flags are used to indicate a particular state of connection such as SYN, ACK, etc. & 0x00\\ \cline{1-3}

Retrans.  &  Indicates if the current record is from a retransmitted packet, caused due to attack or network congestion.& 0\\\cline{1-3}

RTT & Indicator of propagation and processing delay. High RTT can be caused due to MiTM attack.& -1\\\cline{1-3}

Flow Cnt &  Indicates the number of TCP flows in a specific time interval. Indicates the connected and disconnected DNP3 masters. Flow is collection of packets. & -1 \\\cline{1-3}

Flow Fin Cnt &  Indicates if the current flow carries the final packet. & -1\\\cline{1-3}

Packets & Number of packets transmitted in a specific time interval. & -1\\ \cline{1-3}

Snort Alert &  Boolean indicating an alert from snort. & 0\\\cline{1-3}

Alert Type  &  Indicates the alert type such as DNP3, ARP spoof, ICMP flood or any other types. & Nan\\\cline{1-3}

LL Src & Source id of the DNP3 master or outstation. Indicator of which outstation communicates with the master in that specific record. & -1 \\ \cline{1-3}

LL Dest &  Destination id of the DNP3 master or outstation. Indicator of which outstation communicates with the master in that specific record. & -1\\ \cline{1-3}

LL Len  &  Indicator of the DNP3 payload size as well as the function type. Usually the response carries DNP3 point information, hence this length correlates with the function code as well as the outstation currently communicating. & 0\\\cline{1-3}

LL Ctrl & This indicates the initiator of the communication. Determines the primary/secondary server. & 0x00\\\cline{1-3}

TL Ctrl &  Indicates the FIN/FIR/Sequence number for determining if the DNP3 payload is the first or final segment. & 0x00  \\\cline{1-3}

Func. code &  Indicates the function code: either READ, WRITE, OPERATE, DIRECT OPERATE, etc. & -1\\\cline{1-3}

AL Ctrl & Indicates the FIN/FIR/Seq/Confirm and Unsolicited flags. This indicates if there are unsolicited, first, final from application layer standpoint.
& 0x00\\\cline{1-3}

Obj count &  This count determines the number of BI, BO, AI, AO points associated with a substation.& 0\\\cline{1-3}
         
AL Payload &  Contains the DNP3 points used to extract the physical features such as branch status, real power flows and injections in branch and buses for a substation. & Nan\\\cline{1-3}

\end{tabular}
\label{table:fusion_features}
\vspace{-0.5cm}
\end{table}

%% file: Content/6_fusion_types.tex
\section{Fusion}\label{fusion_types}
As presented in Fig.~\ref{fig:architecture_types}, the \textit{Fusion} block involves different types of fusion. Intra-domain and inter-domain are considered for training the IDS using supervised and unsupervised learning techniques. We also explore location-based fusion and visualization for causal inference of the impact of the intrusion in different locations of the network. Finally, co-training with feature split is used to train the IDS using semi-supervised learning with labeled and unlabeled data.

\subsection{Intra-Domain and Inter-Domain Fusion}
Fusion of cyber sensor information from different sources is homogeneous source fusion. For example, the operation of fusing Elasticsearch logs with pyshark or raw packet capture to form the $cyber\_table$ is intra-domain fusion.

Fusion of cyber and physical sensor information from different sources is heterogeneous source fusion. For example, the operation of fusing $cyber\_table$ with $physical\_table$ is inter-domain fusion.  

\subsection{Location-Based Fusion}
In multi-sensor data fusion, sensor location plays a major role. For example, the military uses location-based multi-sensor fusion to estimate the location of enemy troops by amalgamating sensor information from multiple radars and submarines. The challenges associated with different locations stem from time differences in event recognition. A radar can pick up a signal with a different latency than a submarine due to the difference in communication medium as well as its location relative to the enemy troop. Similarly, our sensors such as IDS, firewall alerts, and network logs are positioned at different locations in the network. It is essential to correlate events among different locations before merging them for inferring any attacks.
\begin{figure*} [!htb]
    \centering
    \vspace{-0.2cm}
    \includegraphics[width=1.0\linewidth]{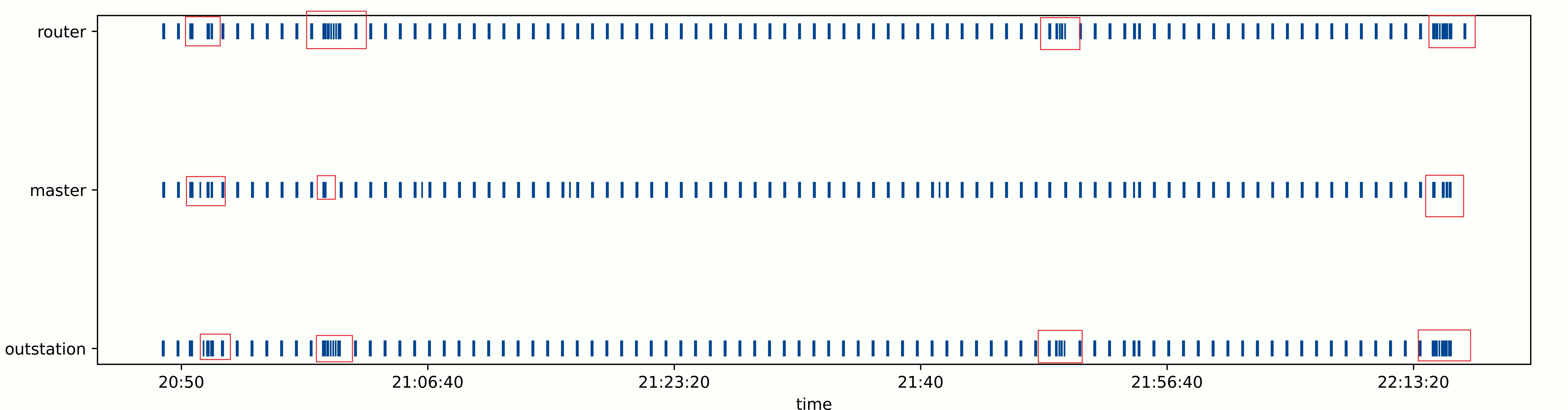}
    \vspace{-0.35cm}
    \caption{Location based fusion from the master, outstation, and substation router. The high density traffic observed in the places marked with red rectangles is an indicator of DoS attack. This fusion assists in causal analysis for determining the initial victim of the DoS intrusion as well as inferring the pattern of impact across other devices in the network.}
    \label{fig:location_fusion}
\end{figure*}

\subsection{Co-Training Based Split and Fusion}
There exist scenarios where labels cannot be captured.  
The co-training algorithm~\cite{co_training} uses feature split when learning from a dataset containing a mix of labeled and unlabeled data. This algorithm is usually preferred for datasets that have a natural separation of features into disjoint sets~\cite{cotraining2}. Since the cyber and physical features are disjoint, we adopt feature split based co-training. 
The approach is to incrementally build classifiers over each of the split feature sets. In our case, we split the fused features into cyber and physical features. Each classifier, $cy\_cfr$ (first 17 features in Table~\ref{table:fusion_features}) and $phy\_cfr$(last 9 features in Table~\ref{table:fusion_features}), is initialized using a few labeled records. At every loop of co-training, each classifier chooses one unlabeled record per class to add to the labeled set. The record is selected based on the highest classification confidence, as provided by the underlying classifier.  Further, each classifier rebuilds from the augmented labeled set, and the process repeats. Finally, the two classifiers $cy\_cfr$ and  $phy\_cfr$ obtained from the co-training algorithm gives probability score against the classes for each record, which is added and normalized to determine the final class of the record~\cite{cotraining2}. The classifiers selected in our experiments are Linear Support Vector Machine (SVM), Logistic Regression, Decision Tree, Random Forest, Naive Bayes, and Multi-Layer Perceptron. 

\begin{figure} [h]
    \centering
    \vspace{-0.2cm}
    \includegraphics[width=1.0\linewidth]{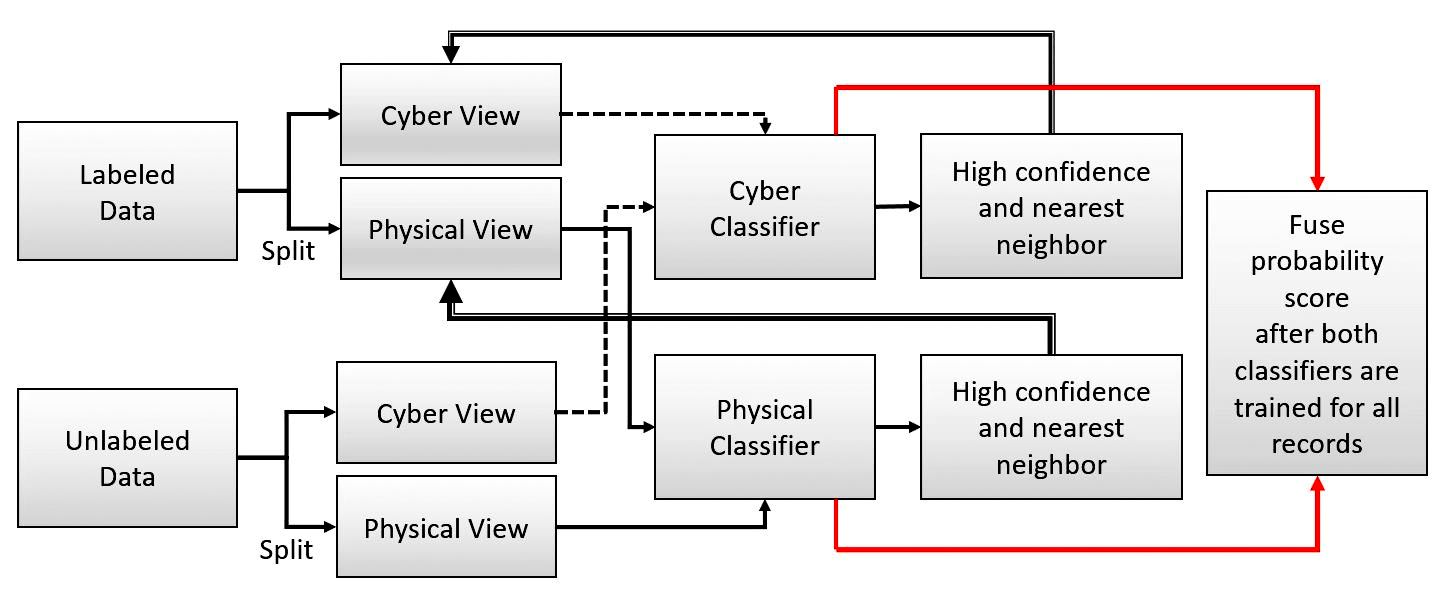}
    \vspace{-0.35cm}
    \caption{Co-training based fusion for  labeled and unlabeled datasets. The fused dataset is split into cyber and physical views and trained in the cyber and physical classifiers separately, finally fusing and normalizing the probability scores for final classification.}
    \label{fig:cotraining}
\end{figure}

%% file: Content/5_data_transformation.tex
\section{Data Transformation}\label{data_transformation}
Real-time testbed data is usually insufficient, conflicting, diverse format and at times lack in certain pattern or trends. Hence, data pre-processing is essential in transforming raw data into an understandable format. The raw data extracted from multiple-sensors are processed through three steps: a) data imputation, b) data encoding, c) data scaling, and d) feature reduction.

\subsection{Data Imputation}

Imputation is a statistical method of replacing the missing data with substituted values. Substitution of a data point is unit imputation, and substituting a component is item imputation. Imputation tries to preserves all the records in the data table by replacing missing data with an estimated value based on other available information or feeds from domain experts.  There are other forms of imputation such as mean, stochastic, regression imputation etc. Imputation can introduce a substantial amount of bias and can also impact efficiency. In this work, we have not tried to address such discrepancies of bias introduced due to imputation.
Since we merge data from different sources with unique features, the chances of missing data are high. Hence, we perform unit imputation in our dataset based on the default values in the $Def$ column of the Table~\ref{table:fusion_features}.

\subsection{Data Encoding}
There are numerous features in the fused dataset which are categorical. These categorical features are encoded using the preprocessing libraries in scikit learn, so that the predictive model can better understand the data. There are different types of encoders such as an ordinal encoder, label encoder, one hot encoder, etc. In this work, we use label encoding. Label encoding is preferred over one hot encoding when the cardinality of the categories in the categorical feature is quite large as it results in the issue of high dimensions. 
We also do not consider an ordinal encoder, as it is processed on the 2D dataset ($samples$*$features$). Since we process cross domain features, we perform encoding on individual features separately using label encoding. 

\subsection{Scaling and Normalization}
Scaling and normalizing the feature is essential for various ML and DL techniques such as Principal Component Analysis (PCA), Multi-Layer Perceptrons (MLPs), Support Vector Machines (SVMs),  etc. Though certain techniques such as Decision Trees or Random Forest, are scale-invariant, it is still essential to normalize and train. Before performing normalization, we perform log transformation and categorical encoding for the features with high variance and varied range of values, respectively. Hence, 
we evaluate both log transformation as well as scaling. Additionally, we considered \textit{Min-Max scaling} as performed in our prior works on intrusion detection on KDD and CIDDS datasets~\cite{ieee_cqr}.

\subsection{Feature Reduction}
\begin{figure} [h]
    \centering
    \vspace{-0.2cm}
    \includegraphics[width=1.0\linewidth]{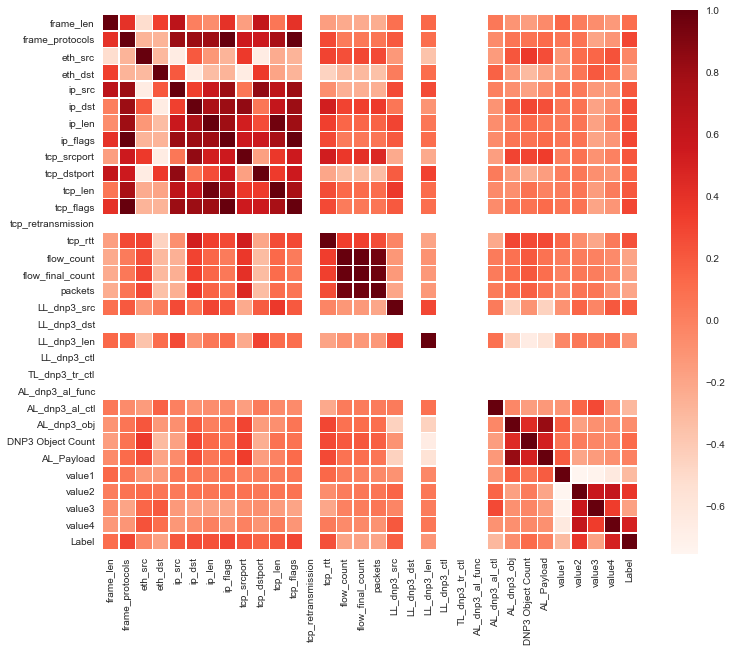}
    \vspace{-0.35cm}
    \caption{Inter-feature correlation based on Pearson Coefficient}
    \label{fig:feature_correlation}
\end{figure}

Once the features from multiple sensors are merged, dimension reduction (inter-feature correlation) is performed to remove the trivial features using Principal Component Analysis (PCA). PCA is a linear dimensionality reduction method that uses Singular Value Decomposition (SVD) on the data to project it to a lower dimensional space \cite{wold1987PCA}. The inter-feature correlation for our fused dataset from RESLab is based on the Pearson Coefficient \cite{pearson}, shown in as shown in Fig.~\ref{fig:feature_correlation}, where it can be observed that intra-domain features have higher correlation amongst each other. There is also some correlation observed across the cyber and physical features. Features with higher correlation are more linearly dependent and thus have a similar effect on dependent variables. For example, if two features have high correlation, one of the two features can be eliminated.

%% file: Content/7_ids.tex
\section{Intrusion Detection Post Fusion}\label{ids}
After the features are extracted, merged, and pre-processed we design IDS using different ML techniques. We have con- sidered manifold learning and clustering as the unsupervised learning techniques, a few linear and non-linear supervised learning techniques, and co-training based semi-supervised learning methods for training the IDS. In this section, we briefly explain the ML techniques we use. 

\subsection{Manifold Learning}
PCA for feature reduction does not perform well when there are nonlinear relationships within the features. Manifold learning is adopted in the scenarios where the projected data in the low dimensional planar surface is not well represented and needs more complex surfaces. Multi-featured data are described as a function of a few underlying latent parameters. Hence the data points can be assumed to be samples from a low-dimensional manifold embedded in a high-dimensional space. These algorithms tries to decipher these latent parameters for low-dimensional representation of the data. There are a lot of approaches to solve this problem such as  Locally Linear Embedding, Spectral Embedding, Multi Dimensional Scaling, IsoMap etc. 

\subsubsection{Locally Linear Embedding (LLE)}
LLE computes the lower-dimensional projection of the high dimensional data by preserving distances within local neighborhoods. It is equivalent to a series of local PCA which are globally compared to obtain the best non-linear embedding~\cite{scikit_manifold}. The LLE algorithm consists of 3 steps~\cite{lle}: a) Compute k-nearest neighbor for a data point. b) Construct a weight matrix associated with the neighborhood of each data point. Obtains the weights 
that best reconstruct each data from its neighbors, minimizing the cost.
c) Compute the transformed data point $Y$ best reconstructed by the weights, minimizing the quadratic form.



\subsubsection{Spectral Embedding}
Spectral embedding builds a graph incorporating neighborhood information. Considering the Laplacian of the graph, it computes a low dimensional representation of the data set that optimally preserves local neighborhood information~\cite{se}. Minimization of a cost function, based on the graph ensures that points closer on the manifold are mapped closer in the low dimensional space, preserving local distances~\cite{scikit_manifold}. 
The Spectral Embedding algorithm consists of 3 steps: a) Weighted Graph Construction in which raw data are input into a graph representation using an adjacency matrix. b) Construction of unnormalized and a normalzied graph Laplacians as $L = D- A$ and $L = D^{-0.5}(D-A)D^{-0.5}$, respectively. c) Finally, partial eigenvalue decomposition is done on the graph Laplacian.

\subsubsection{Multi Dimensional Scaling (MDS)}
MDS performs projection to lower dimension to improve interpretability while preserving `dissimilarity’ between the samples. It preserves the dissimilarity by minimizing the square difference of the pairwise distances between all the training data between the projected, lower dimensional and the original higher dimensional space, 
\begin{equation}
    \operatorname{Diff}_{P}\left(X_{1}, \ldots, X_{n}\right)=\left(\sum_{i=1}^{n} \sum_{j=1 \mid i \neq j}^{n}\left(\left\|x_{i}-x_{j}\right\|-\delta_{i, j}\right)^{2}\right)^{1 / 2}
\end{equation}

\noindent where $\delta_{i,j}$ is the general dissimilarity metric in the original higher dimensional space and $\left\|x_{i}-x_{j}\right\|$ is the projected/lower dimensional dissimilarity pairwise between training samples $i$ and $j$. The model can be finally validated by a scatter plot of pairwise distance in projected and original space. There are two types of MDS: Metric and Non-Metric based.  In Metric MDS, the distances between the two points in projection are set to be as close as possible to the dissimilarity (or distance) in original space. Non-metric MDS tries to preserve the order of the distances, and hence seeks a monotonic relationship between the distances in the embedded  and original space.


\subsubsection{t-SNE Visualization}
The manifold learning technique called t-distributed Stochastic Neighbor Embedding is useful to visualize high-dimensional data, as it reduces the tendency of points to crowd together at the center. This technique converts similarities between data records to joint probabilities and then tries to minimize the Kullback-Leibler divergence (technique used to compare two probability distributions) between the joint probabilities of the low-dimensional embedding and the high-dimensional data using gradient descent. The only issue with the use of this technique is that it is computationally expensive and is limited by two or three embeddings in some methods. In our intrusion detection methods, our purpose is to evaluate if in the low-dimensional embedding we can find some correlation of the data points with the labels.

\subsubsection{IsoMap Embedding}
Isomap stands for isometric mapping and is an extension to the MDS technique discussed earlier. It uses geodesic
paths instead of eucledian distance for nonlinear dimensionality reduction. Since MDS tries to preserve large pairwise distance over the small pairwise distance, Isomap first determine a neighborhood graph by finding the k nearest neighbor of each point, further connecting this points in the graph and assigns weights. Then it computes the shortest geodesic path between all pairs of point in the graph, to use this distance measure between connected points as weights to apply MDS to the shortest-path distance matrix~\cite{extra_fusion_ref1}. 

\subsection{Clustering}
One of the fundamental problems in multi-sensor data fusion is \textit{data association}, where different observations in the dataset are grouped into clusters~\cite{hall_book}. Hence, various clustering techniques are explored for data association.  
\subsubsection{K-means Clustering}
The k-means algorithm clusters data by separating samples in $n$ groups of equal variance, minimizing a criterion known as the inertia. The algorithm starts with a group of randomly selected centroids, which are used as the beginning points for every cluster, then performs iterative calculations to optimize the positions of the centroids by minimizing inertia. The process stops when either 
the centroids have stabilized or the number of iterations has been achieved.

\subsubsection{Spectral Clustering}
The main concept behind spectral clustering is the graph Laplacian matrix. The algorithm takes the following steps~\cite{spectral_clustering}:
\begin{enumerate}
    \item Construct a similarity graph either based on an $\epsilon$-neighborhood graph, a $k-nearest$ neighbor graph, or a fully connected graph.
    \item Compute the normalized Laplacian $L$.
    \item Compute the first $k$ eigen-vectors $u_{1},u_{2}...,u_{k}$ of $L$. The first eigen-vectors are related to the $k$ smallest eigen values of $L$.
    \item Let $U \in R^{n*k}$ be the matrix containing the vectors $u_{1},u_{2}...,u_{k}$ as columns.
    \item For $i=1,,,,n$, let $y_{i} \in R^{k}$ be the vector corresponding to the $i^{th}$ row of $U$.
    \item Cluster points $(y_{i})$ in $R^{k}$ with k-means algorithm into clusters $C_{1},...C_{k}$.
\end{enumerate}

\subsubsection{Agglomerative Clustering}
Agglomerative clustering in a bottom-up manner, where at the beginning, where each object belongs to one single-element cluster, which are the leaf clusters of a dendogram. At each step of the algorithm, the two clusters that are most similar (based on a similarity metric such as distance) are combined into a larger cluster. The procedure is followed until all points are members of a single big cluster. The steps form a hierarchical tree, where a distance threshold is used considered to cut the tree to partition the data into clusters. As per scikit, this algorithm recursively merges the pair of clusters that minimally increases a given linkage distance~\cite{scikit_clustering}. The parameter \textit{distance\_threshold} in the scikit-learn implementation is used to cut the dendogram.

\subsubsection{Birch Clustering}
The Balanced Iterative Reducing and Clustering Using Hierarchies (BIRCH) \cite{zhang1996birch} algorithm is more suitable for the cases where the amount of data is large and the number of categories K is also relatively large. It runs very fast, and it only needs a single pass to scan the data set for clustering.


\subsection{Supervised Learning}
Though manifold learning and clustering techniques helps visualize and cluster the data samples in the intrusion time-interval from the non-intrusion ones, still the results of these techniques are hard to validate without any labels, hence various supervised learning techniques are also considered in designing the anomaly based IDS.

\subsubsection{Support Vector Classifier (SVC)}
 Support vector machine builds an hyperplane or set of hyperplanes in a higher dimensional space which are further used as a decision surface for classification or outlier detection. It is a supervised learning based classifier  which performs better even for scenarios with higher feature size than sample size. The decision function, or support vectors, defined using the  kernel type such as sigmoid, polynomial, linear or radial basis function plays a major impact on the classifier performance. Different variants of SVCs have been predominantly proposed in intrusion detection solutions~\cite{svm1,svm2}.


\subsubsection{Logistic Regression (LR) Classifier}
LR is a classification algorithm, used mainly for discrete set of classes. It is a probability-based classification technique which minimizes the error cost using the logistic sigmoid function. It uses the gradient descent technique to reduce the error cost function. Industries make a wide use of it, since it is very efficient and highly interpretable\cite{lr}.


\subsubsection{Naive Bayes (NB) Classifier}
NB is a supervised learning technique based on the Bayes Theorem, with the naive assumption of independent features, conditioned on the class. Based on the feature likelihood distribution, they posses different forms: Gaussian, Bernoulli, Categorical, Complement, etc. Though it is computationally efficient, the selection of feature likelihood may alter results. Spam filtering, text classification, and also in network intrusion detection it is used profusely~\cite{nb1}. A naive-bayes based solution was proposed for IDS in a smart meter network~\cite{sahu_naive_bayes}.

\subsubsection{Decision Tree (DT) Classifier}
The advantage of using DT is that it requires the least data transformation. Fundamentally it creates internally, models that predicts the target class by learning decision rules inferred from the features. 
This technique sometimes meet with over-fitting issues while learning complex trees that are hard to generalize. Hence, it adopts pruning techniques such as reducing the tree max-depth to deal with over-fitting. If data in the samples are biased it may highly likely create biased trees. The computation cost of using this classifier is logarithmic in the number of data records. It has been used in protocol classification problem~\cite{dt3,dt4} for classifying anomalous packets.



\subsubsection{Random Forest (RF) Classifier}
Basically, RF creates decision trees on randomly picked data samples, further computes prediction from each tree and selects the best solution through voting. More trees results in a more robust forest. It is an ensemble based classifier in which a diverse collection of classifiers (decision trees) are constructed by incorporating randomness in tree construction.
Randomness decreases the variance to address the overfit issues prevailing in DT. While comparing with SVMs, RF is fast and works well with a mixture of numerical and categorical features.
It has a variety of applications, such as recommendation engines, image classification and feature selection.
Due to its variance reduction feature and least need of data pre-processing, it is also preferred in the cyber security area~\cite{rf1,rf2}.

\subsubsection{Neural Network (NN) Classifier}
Neural networks is effective in the case of complex non-linear models. In our IDS classification problem, we make use of multi-layer perceptron (MLP) as the supervised learning algorithm. It learns a non-linear function approximator whose inputs are the features for a record and outputs the class. Unlike a logistic regressor, it comprises of multiple hidden layers. A major issue with NN models is it large set of hyper tuning parameter such as number of hidden neurons, layers, iterations, dropouts, etc., that can affect the hyper-parameter tuning process for improving accuracy. Additionally, it is quite sensitive to feature scaling. 
Following Occam's razor, security professionals tend to avoid 
neural networks in intrusion detection, wherever possible. Still NN can be explored to capture temporal pattern with the use of Recurrent Neural Networks (RNN) and spatial pattern using Graph Neural Networks (GNN). 


%% file: Content/8_results.tex
\section{Results and Analysis}\label{results}


In this section, we study the improvement of the detection performance of IDS, when a fused dataset is considered in comparison to the use of only cyber or physical features. We design the IDS as a classifier when training with supervised and semi-supervised based ML techniques. We analyze the IDS performance based on the different types of MiTM attack carried out in the RESLab testbed. For supervised learning techniques, we analyse the impact of labeling as well as feature reduction on the detection accuracy. For unsupervised learning techniques, we compare the performance of the clustering techniques based on different metrics. In most of the experiments, we expect to receive highest scores for 
either 2 or 3 clusters, since we want to cluster attacked traffic from non-attacked. The third cluster can be an undetermined cluster. We also utilize and test a co-training based semi-supervised learning technique by assuming loss of labels for some experiments and compare them with supervised learning techniques. 


\subsection{Supervised Technique Intrusion Detection with Snort Alert as Label}

\subsubsection{Metrics for evaluation}
The IDS performance is evaluated by classifier's accuracy computed using metrics such as \textit{Recall}, \textit{Precision}, and \textit{F1-score}. Recall is the ratio of the true positives to the sum of true positives and false negatives. Precision is the ratio of the true positives to the sum of true positives and false positives.
High precision is ensured by a low false positive rate. High recall is an indication of low false negative rate. 
False negatives are highly unwanted in security, since an undetected attack may result in more privilege escalations and can impact a larger part of network. False positives is expensive as time and money is invested for security professionals to investigate a non-critical alert.
Hence, harmonic mean of recall and precision, called F1-score, is a preferred metric for a balanced evaluation.

\subsubsection{Labels Evaluation}
The performances are compared, considering labels from Snort alerts and labels based on the intruders' attack windows, to train the supervised learning based IDS classifiers. The intruders' attack window is the difference between the attack script end and start time. We label every record in this window belonging to the compromised class. It is interesting to observe from Table~\ref{tab:label_comparison} that the classifier trained using the attack window label performed better than the Snort labels, based on the average F1-score, Recall, and Precision. 
These metrics are computed by taking the average of all the metrics from different use cases. This analysis indicates that training a model from well-known IDS may not act as an ideal classifier for intrusion detection. Hence, for our further studies, we train the classifier using the attack window based label. 

\begin{table}[h]
    \centering
    \begin{tabular}{|l|l|l|l|l|l|l|}
     \hline
Classifier & \multicolumn{3}{|c|}{Snort Label} & \multicolumn{3}{|c|}{Label from Attack Window}\\
    \hline
        Avg. & F1 score & Rec. & Prec. & F1 score & Rec. & Prec.\\ \hline
        SVC & .566 & .69 & .496 & .752 & .776 & .799 \\ \hline
        DT & .738 & .73 & .757 & .909 & .909 & .92 \\ \hline
        RF & .764 & .789 & .776 & .891 & .896 & .903 \\ \hline
        GNB & .598 & .574 & .745 & .724 & .729 & .748 \\ \hline
        BNB & .57 & .589 & .621 & .634 & .655 & .676 \\ \hline
        MLP & .561 & .671 & .491 & .621 & .695 & .604 \\ \hline
    \end{tabular}
    \caption{Comparison of the labels using different classifier based on the evaluation metrics.}
    \label{tab:label_comparison}
    \vspace{-0.2 cm}
\end{table}

\subsubsection{Use Case Specific Evaluation} 
We analyze the dataset constructed from four use cases based on different strategies of FDI and FCI attacks (measurement and control, respectively). These cases use different polling rates and DNP3 masters on the synthetic 2000-bus grid case illustrated in the RESLab paper~\cite{Sahu2020}. Use Case 1 and 2 are FCI attacks on binary and mixed binary/analog commands from the control center to some selected outstations, selected from our prior work on graph-based contingency discovery~\cite{n_x}. Use Case 3 and 4 are a mix of FCI and FDI attacks. These use cases differ based on the type and sequence of modifications done by the intruder as shown in Table~\ref{tab:uc_table}.

\begin{table*}[t]
    \centering
    \begin{tabular}{|l|l|l|l|}
     \hline
 \multicolumn{2}{|c|}{FCI} & \multicolumn{2}{|c|}{FCI with FDI}\\
    \hline
         UC1 & UC2 & UC3 & UC4\\ \hline
        Binary Commands & Analog, Binary Commands & Measurements=$>$Commands & Measurements=$>$ Commands=$>$Measurements \\ \hline
    \end{tabular}
    \caption{Use cases based on the type and sequence of modifications.}
    \label{tab:uc_table}
    \vspace{-0.2 cm}
\end{table*}

Due to the variation of attempts an intruder needs to take to implement the use cases, the number of samples collected for every scenario differs. 
In the MLP based classifier, the number of samples plays a vital role; hence, MLP performs better for scenarios with the number of DNP3 masters equal to 10 versus 5 and with a DNP3 polling interval of 30 s versus 60 s. The DT and RF classifiers outperform the other classifiers in almost all the scenarios. The NB classifiers, both Gaussian and Bernoulli, need the features to be independent for optimal performance. Since most of the features are strongly correlated based on Fig.~\ref{fig:feature_correlation}, the performance of NB is relatively weak compared to other classifiers. Usually, Gaussian Naive Bayes (GNB) is considered for features that are continuous and Bernoulli Naive Bayes (BNB) for discrete features. In our fused dataset, since we have both types of features, we consider both techniques for evaluation. In majority of the scenarios, GNB performed better than BNB, indicating the physical features have more impact on the detection compared to categorical cyber features. Table~\ref{tab:uc_comparison} shows the comparison of classifiers for different use cases, and Table~\ref{tab:uc_comparison_gridsearch} shows the comparison using grid search cross validation based tuning of hyper-parameters for each classifier.

\begin{table}[!h]
    \begin{tabular}{|l|l|l|l|l|l|l|l|l|l|}
     \cline{1-9}
\multicolumn{3}{|c|}{Scenarios} & \multicolumn{6}{|c|}{Classifiers}\\
    \cline{1-9}
    uc & masters & PI & SVC & DT & RF & GNB & BNB & MLP\\
    \cline{1-9}
        \multirow{2}{*}{\textbf{UC1}}  & 10 & 30 & .70 & .74 & .75 & .59 & .70 & .70\\ \cline{2-9}
        & 10 & 60 & .78 & .87 & .81 & .75 & .49 & .58\\ \cline{1-9}
        \multirow{4}{*}{\textbf{UC2}}  & 5 & 30 & .88 & .76 & .92 & .73 & .52 & .86\\ \cline{2-9}
       & 5 & 60 & .88 & .89 & 1.0 & .94 & .89 & .66\\ \cline{2-9}
        & 10 & 30 & .84 & .93 & .93 & .73 & .59 & .77\\ \cline{2-9}
       & 10 & 60 & .64 & .97 & .88 & .33 & .58 & .52\\ \cline{1-9}
       \multirow{4}{*}{\textbf{UC3}}  & 5 & 30 & .95 & .98 & .93 & .93 & .57 & .72\\ \cline{2-9}
       & 5 & 60 & .50 & 1.0 & .88 & .72 & .33 & .40\\ \cline{2-9}
       & 10 & 30 & .85 & 1.0 & .97 & .83 & .66 & .86\\ \cline{2-9}
       & 10 & 60 & .89 & .98 & .91 & .84 & .73 & .91\\ \cline{1-9}
       \multirow{4}{*}{\textbf{UC4}}  & 5 & 30 & .59 & .86 & .88 & .56 & .54 & .39\\ \cline{2-9}
       & 5 & 60 & .63 & .81 & .77 & .74 & .77 & .31\\ \cline{2-9}
       & 10 & 30 & .65 & .96 & .97 & .63 & .78 & .57\\ \cline{2-9}
       & 10 & 60 & .75 & .98 & .88 & .83 & .80 & .50\\ \cline{1-9}
       \hline
    \end{tabular}
    \caption{Comparison of the classifier based on the scenarios i.e. use cases, number of masters and the polling interval (PI) in sec. }
    \label{tab:uc_comparison}
    \vspace{-0.2 cm}
\end{table}

\begin{table}[!h]
    \begin{tabular}{|l|l|l|l|l|l|l|l|l|l|}
     \cline{1-9}
\multicolumn{3}{|c|}{Scenarios} & \multicolumn{6}{|c|}{Classifiers}\\
    \cline{1-9}
    uc & masters & PI & SVC & DT & RF & GNB & BNB & MLP\\
    \cline{1-9}
        \multirow{2}{*}{\textbf{UC1}}  & 10 & 30 & .70 & .78 & .75 & .70 & .69 & .70\\ \cline{2-9}
        & 10 & 60 & .54 & .87 & .81 & .78 & .52 & .7\\ \cline{1-9}
        \multirow{4}{*}{\textbf{UC2}}  & 5 & 30 & .51 & .88 & .84 & .72 & .51 & .67\\ \cline{2-9}
       & 5 & 60 & .66 & 1.0 & 1.0 & .83 & .89 & .62\\ \cline{2-9}
        & 10 & 30 & .45 & .94 & .89 & .81 & .44 & .86\\ \cline{2-9}
       & 10 & 60 & .52 & .97 & .85 & .75 & .61 & .58\\ \cline{1-9}
       \multirow{4}{*}{\textbf{UC3}}  & 5 & 30 & .36 & .98 & .93 & .93 & .50 & .91\\ \cline{2-9}
       & 5 & 60 & .40 & 1.0 & .96 & .88 & .26 & .44\\ \cline{2-9}
       & 10 & 30 & .41 & 1.0 & .99 & .84 & .63 & .69\\ \cline{2-9}
       & 10 & 60 & .40 & .93 & .88 & .89 & .76 & .82\\ \cline{1-9}
       \multirow{4}{*}{\textbf{UC4}}  & 5 & 30 & .39 & .97 & .93 & .57 & .56 & .61\\ \cline{2-9}
       & 5 & 60 & .31 & .63 & .68 & .65 & .77 & .68\\ \cline{2-9}
       & 10 & 30 & .44 & .96 & .95 & .65 & .78 & .65\\ \cline{2-9}
       & 10 & 60 & .50 & .88 & .85 & .80 & .80 & .50\\ \cline{1-9}
       \hline
    \end{tabular}
    \caption{\textbf{Optimal HyperParameter with GridSearch} Comparison of the classifier based on the scenarios i.e. use cases, number of masters and the polling interval (PI) in sec. }
    \label{tab:uc_comparison_gridsearch}
    \vspace{-0.2 cm}
\end{table}

\subsubsection{Impact of Fusion}
We evaluate the performance of the classifier by considering pure physical and pure cyber based intra-domain fusion as well as cyber-physical inter-domain fusion. The pure physical and cyber physical based fusion outperforms pure-cyber based fusion for all the classifiers shown in Table~\ref{tab:fusion_type_comparison}. Hence, it indicates that introduction of physical side features can improve the accuracy of conventional IDS that only considers network logs in communication domain. The pure physical features relatively performed better than cyber physical because in the testbed, only few features (i.e. measurements for the impacted substation) are considered for extraction. If we consider all the measurements from the grid simulation, the detection accuracy will decrease due to feature explosion. Feature reduction techniques  such as PCA for the physical features  may not be an ideal solution for a huge synthetic grid. 

\begin{table}[!h]
    \centering
    \begin{tabular}{|l|l|l|l|l|l|l|l|l|l|}
     \hline
Clfr & \multicolumn{3}{|c|}{Pure Cyber} & \multicolumn{3}{|c|}{Pure Physical} & \multicolumn{3}{|c|}{Cyber Physical}\\
    \hline
         Avg. & F1 & Rec. & Pre. & F1 & Rec. & Pre. & F1  & Rec. & Pre.\\ \hline
        SVC & .62 & .68 & .59 & .75 & .77 & .80 & .75 & .77 & .80\\ \hline
        DT & .77 & .77 & .77 & .93 & .93 & .94 & .91 & .91 & .92\\ \hline
        RF & .69 & .69 & .68 & .92 & .92 & .93 & .89 & .90 & .90\\ \hline
        GNB & .58 & .57 & .59 & .78 & .77 & .81 & .72 & .73 & .75\\ \hline
        BNB & .52 & .56 & .55 & .65 & .68 & .66 & .63 & .66 & .68\\ \hline
        MLP & .56 & .66 & .53 & .72 & .76 & .77 & .62 & .70 & .61\\ \hline
    \end{tabular}
    \caption{Comparison of the classifier with pure cyber fusion, pure physical fusion, and cyber-physical fusion features }
    \label{tab:fusion_type_comparison}
    \vspace{-0.2 cm}
\end{table}

\subsubsection{Impact of Feature Reduction}
In this subsection, we analyze feature reduction techniques such as PCA  and Shapiro ranking for feature reduction and feature filtering to evaluate the performance of the IDS. Table~\ref{tab:feature_selection_comparison} illustrates the performance scores for different classifiers with PCA transformed features and shapiro features selected for scores more than 0.7.   It can be observed that except for the DT and RF, other classifier's performance improved by both operation. DT and RF behaves the best when most of the features are kept intact. In most of the case selection of features based on Shapiro features performed better than PCA transformation. Still the total variance threshold taken may impact the number of principal components considered, which can affect the results.


\begin{figure} [!h]
    \centering
    \vspace{-0.2cm}
    \includegraphics[width=1.0\linewidth]{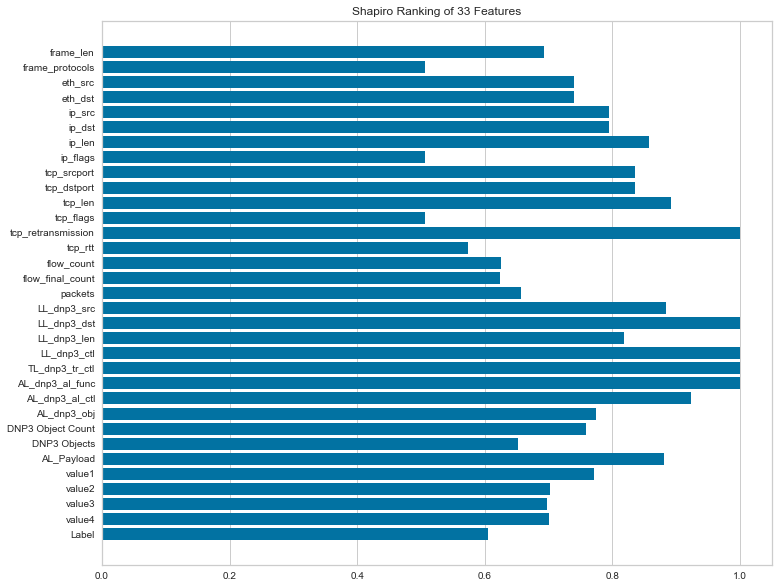}
    \vspace{-0.35cm}
    \caption{Ranking feature importance for extracting features. Of all the features, scores above 0.7 is selected for training.}
    \label{fig:shapiro_ranking}
\end{figure}

\begin{table}[h]
    \centering
    \begin{tabular}{|l|l|l|l|l|l|l|l|l|l|}
     \hline
Clfr & \multicolumn{3}{|c|}{All Features} & \multicolumn{3}{|c|}{PCA} & \multicolumn{3}{|c|}{Shapiro Ftrs $\ge$ 0.7}\\\hline
         Avg. & F1 & Rec. & Pre. & F1 & Rec. & Pre. & F1  & Rec. & Pre.\\ \hline
        SVC & .75 & .77 & .80 & .77 & .80 & .81 & .77 & .78 & .79\\ \hline
        DT & .91 & .91 & .92 & .82 & .82 & .83 & .89 & .89 & .91\\ \hline
        RF & .89 & .90 & .90 & .86 & .86 & .87 & .84 & .84 & .84\\ \hline
        GNB & .72 & .73 & .75 & .77 & .78 & .78 & .83 & .84 & .87\\ \hline
        BNB & .63 & .66 & .68 & .74 & .76 & .76 & .80 & .82 & .86\\ \hline
        MLP & .62 & .70 & .61 & .61 & .68 & .64 & .50 & .64 & .41\\ \hline
    \end{tabular}
    \caption{Comparison of the classifier with all features, reduced feature with PCA transformation, and feature selection based on shapiro ranking }
    \label{tab:feature_selection_comparison}
    \vspace{-0.2 cm}
\end{table}



\subsection{Unsupervised Learning Techniques}

\subsubsection{Metrics for evaluation}
For evaluating the performance of the clustering techniques, the Silhoutte scores, Calinski Harabasz score, Adjusted Rand score, and Davies Bouldin scores are considered. The \textit{Silhoutte score} (S) is the mean Silhouette Coefficient of all samples. The Silhouette Coefficient is calculated using the mean intra-cluster distance (a) and the mean nearest-cluster distance (b) for each sample, using $\frac{b - a} {max(a, b)}$. The \textit{Calinski Harabasz score} (CH) is computed based on~\cite{ch_score}. It is the ratio between the within-cluster dispersion and the between-cluster dispersion. The \textit{Rand Index} computes a similarity measure between two clusterings by considering all pairs of samples and counting pairs that are assigned in the same or different clusters in the predicted and true clusterings. This index is further adjusted to be called the Adjusted Rand Index (AR). The \textit{Davies Bouldin score} (DB) is defined as the average similarity measure of each cluster with its most similar cluster, where similarity is the ratio of within-cluster distances to between-cluster distances~\cite{db_score}. Thus, clusters which are farther apart and less dispersed will result in a better score.


\subsubsection{Clustering}
Prior to the clustering techniques, we scaled and normalized the dataset using scaler  and normalize functions since otherwise there will be feature-based bias. We implement four types of clustering techniques: Agglomerative, k-means, Spectral and Birch clustering, to evaluate the optimal number of clusters based on the S, CH, AR, and DB scores. For determining the clusters, we merged the samples from all the use cases to form a larger dataset and then trained the clustering methods by tuning the number of clusters hyper-parameter ($N_c$) from 2 to 10. Fig 8 (a-e)
show the clustered plots using Agglomerative clustering with different number of clusters.
The number of clusters, or centroids, are selected for hyper-parameter tuning since it is found to be the most important factor for success of the algorithm~\cite{clus_ref}.
Ideally, there need to be 3 clusters for un-attacked, attacked with DNP3 alerts, and attacked with ARP alerts, but the distance metric considered results in a greater number of clusters in some methods. Among all the clustering techniques presented in the previous section, the affinity propagation technique does not converge to obtain the exemplars with default paramaters ($damping$ =50, $convergence\_iter$ =200). Hence, the damping and maximum convergence iteration parameters are increased to 0.95 and 2000 respectively, resulting in 34 clusters. The S, CH, DB, and AR scores obtained are 0.605, 3658.1, 0.736, and 0.00085 respectively. 


\begin{table}[h]
    \centering
    \begin{tabular}{|l|l|l|l|l|}
    \hline
         Clustering Algo & S & CH & AR & DB\\ \hline
        Agglomerative & 3 & 3 & 2 & 6\\ \hline
        K-means & 3 & 5 & 2 &  6\\ \hline
        Spectral & 3 & 5 & 2 & 6 \\ \hline
        Birch & 3 & 3 & 3 & 2
        \\ \hline
    \end{tabular}
    \caption{Optimal clusters (Opt $N_c$) using different algorithm obtained using four different evaluation metric}
    \label{tab:unsup_cluster_comparison}
    \vspace{-0.2 cm}
\end{table}

\begin{figure*}[!htb]
  \centering
  \subfigure[]{\includegraphics[height=1.3 in,width=1.37 in]{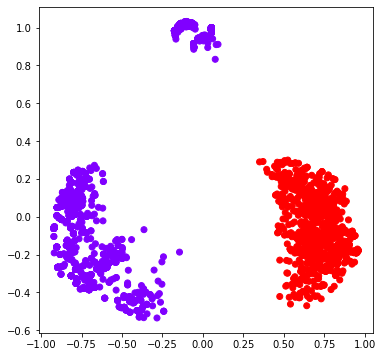}\label{ac2}}
  \subfigure[]{\includegraphics[height=1.3 in,width=1.37 in]{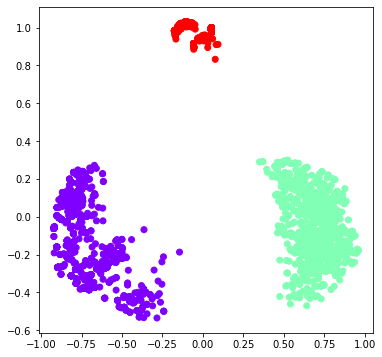}\label{ac3}}
  \subfigure[]{\includegraphics[height=1.3 in,width=1.37 in]{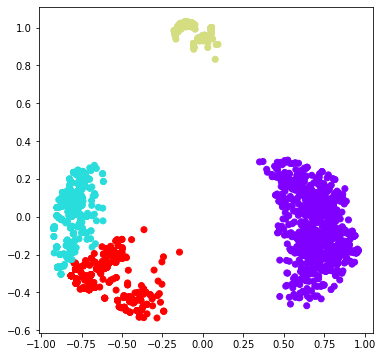}\label{ac4}}
  \subfigure[]{\includegraphics[height=1.3 in,width=1.37 in]{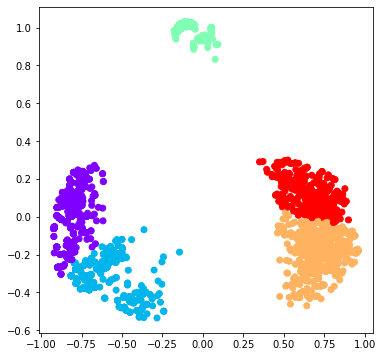}\label{ac5}}
  \subfigure[]{\includegraphics[height=1.3 in,width=1.37 in]{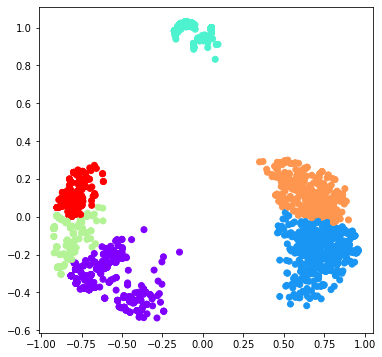}\label{ac6}}
  \caption{Agglomerative clustering with different number of clusters. Clustering with size 2 and 3 outperforms others, validating the detection accuracy of a attacked traffic from a non-attacked one.}
\end{figure*}



\subsubsection{Impact of Fusion}
Considering only physical side fea- tures, most of the evaluation metrics computed very low or negative (in the case of Adjusted Rand index) values, indicating inefficient clusters. The scores of the optimal clusters with combined cyber-physical features had an AR score of more than 0.8, but its maximum is 0.01 for 6 clusters with only physical features. The pure cyber features performed similar to the cyber physical case, but the scores are less compared to the merged features. Hence, it is essential to fuse cyber and physical features prior to perform clustering based unsupervised learning.
\begin{table}[h]
    \centering
    \begin{tabular}{|l|l|l|l|l|l|l|l|l|}
    \hline
    & \multicolumn{4}{|c|}{Pure Cyber} & \multicolumn{4}{|c|}{Pure Physical}\\ \hline
         Clustering  & S & CH & AR & DB& S & CH & AR & DB\\ \hline
        Agglo. & 3 & 5 & 2 & 6 & 2 & 6 & 6 (neg) & 2\\ \hline
        K-means & 3 & 6 & 2 &  5 & 3 & 6 & 6 (neg) & 2\\ \hline
        Spectral & 3 & 5 & 2 & 6 & 3 & 3 & 6 (neg) & 2 \\ \hline
        Birch & 3 & 3 & 2 & 2 & same & 3 & 3 (neg) & 2
        \\ \hline
    \end{tabular}
    \caption{Comparison of Optimal clusters (Opt $N_c$) using different algorithm considering pure cyber and physical features}
    \label{tab:unsup_cluster_cyphy_comparison}
    \vspace{-0.2 cm}
\end{table}

\subsubsection{Robustness}
The robustness of the clustering techniques can be evaluated based on the variance of these evaluation metrics with respect to a) hyper-parameter tuning and b) dataset alterations. In the first case, the mean, variance, and normalized variance ($NVar$ = $\frac{sd}{mean}$) of the evaluation metric $S$, $CH$, $AR$, and $DB$ are computed by altering $N_c$ from 2 to 10 and using the complete dataset extracted for all the use cases. In the second case, similar statistics are computed by keeping the number of clusters fixed at $N_c=3$ and altering the dataset i.e. by using different use cases. A clustering technique that has a lower normalized variance is more robust, and a better mean score is more accurate. Based on the silhoutte scores ($S$) from Table~\ref{tab:clustering_robustness}, k-mean based clustering is found to be more robust to varying data source and has a better mean score, but a main limitation of k-means is its strong dependence on $N_c$. Still, k-means is used in many practical situations such as anomaly detection~\cite{kmean} due to its low computation cost.    

\begin{table}[h]
    \begin{tabular}{|l|l|l|l|l|l|l|l|}
     \cline{1-8}
\multicolumn{2}{|c|}{Scenarios} & \multicolumn{3}{|c|}{Effect of Parameters}& \multicolumn{3}{|c|}{Effect of Data Alt.}\\
    \cline{1-8}
    Met & Algo & Mean & Var & NVar & Mean & Var & NVar\\
    \cline{1-8}
        \multirow{4}{*}{\textbf{S}} 
        & Agg & .52 & .0175 & .254 & .609 & .01 & .164\\ \cline{2-8}
        & K-m & .54 & .013 & .212 & \bftab .615 & .008 & \bftab .145\\ \cline{2-8}
        & Spec & .504 & .021 & .287 & .581 & .015 & .213 \\ \cline{2-8}
         & Bir & \bftab .74 & .011 & \bftab .146 & .599 & .010 & .172 \\ \cline{1-8}
        \multirow{4}{*}{\textbf{CH}}  
       & Agg & 9965 & \num[round-precision=2,round-mode=figures,
     scientific-notation=true]{5516755} & \bftab .235 & 337 & 36880 & .569\\ \cline{2-8}
        & K-m & \bftab 10822 & \num[round-precision=2,round-mode=figures,
     scientific-notation=true]{66329012} & .237 & \bftab 346 & 35690 & .545\\ \cline{2-8}
        & Spec & 8765 & \num[round-precision=2,round-mode=figures,
     scientific-notation=true]{10066904} & .362 & 311 & 34047 & .592 \\ \cline{2-8}
         & Bir & 10484 & \num[round-precision=2,round-mode=figures,
     scientific-notation=true]{13395600} & .349 & 331 & 35637 & .57 \\ \cline{1-8}
       \multirow{4}{*}{\textbf{AR}}  
       & Agg & .703 & .035 & .266 & .534 & .029 & .319\\ \cline{2-8}
        & K-m & .672 & .027 & \bftab .248 & .529 & .022 & \bftab .281\\ \cline{2-8}
        & Spec & \bftab .714 & .039 & .278 & \bftab .638 & .049 & .349 \\ \cline{2-8}
         & Bir & .342 & .014 & .35 & .589 & .047 & .368 \\ \cline{1-8}
       \multirow{4}{*}{\textbf{DB}} 
      & Agg & .026 & 0.0 & \bftab .32 & .053 & .003 & 1.038\\ \cline{2-8}
        & K-m & .54 & 0.0 & .322 & .063 & .003 & .925\\ \cline{2-8}
        & Spec & .504 & 0.0  & .559 & .058 & .003 & 1.037\\ \cline{2-8}
         & Bir & \bftab .74 & 0.0  & 1.344 & \bftab .065 & .003 & \bftab .895\\ \cline{1-8}
       \hline
    \end{tabular}
    \caption{Evaluation of the Robustness of the Clustering Algorithm by varying hyper-parameters and data source.}
    \label{tab:clustering_robustness}
    \vspace{-0.2 cm}
\end{table}

\subsubsection{Manifold Learning}
Manifold learning is adopted for the purpose of visualization. 
%
For quantitative comparisons, we need to employ classification techniques on the features projected in the lower dimensions using these embeddings.  We evaluate the performance of manifold learning methods by testing them with the classifiers presented in the previous subsection.  Table~\ref{tab:manifold_learning_comparison} presents the comparison of the LLE, MDS, spectral, t-SNE, and IsoMap \cite{tenenbaum2000isomap} embeddings considered for classification using SVC, k-NN, DT, RF, GNB, BNB and MLP. Inter-domain fusion doesnt gain much from manifold learning, but an interesting observation is made on the decrease in the difference of F1-scores among the high performing DT and RF classifiers, with the low performing SVC and k-NN classifiers. Hence, we conclude that it is unadvisable to perform manifold learning for our datasets, if training using Decision Tree or Random Forest. The IsoMap embedding that preserves local features of the data by first determining neighbor-hood graph and uses MDS in its last stage performs better than MDS for all the classifier only with the exception of SVC.

\begin{table}[h]
    \centering
    \begin{tabular}{|l|l|l|l|l|l|l|l|}
     \hline
        Clfr$\rightarrow$ & SVC & k-NN & DT & RF & GNB & BNB & MLP\\\hline
        Manifold $\downarrow$ & \multicolumn{7}{|c|}{F1 scores}\\\hline
        LLE & .66 & .74 & .66 & .64 & .38 & .39 & .49\\ \hline
        MDS & .65 & .78 & .77 & .80 & .54 & .48 & .55\\ \hline
        Spectral & .61 & .75 & .73 & .75 & .61 & .62 & .54\\ \hline
        t-SNE & .64 & .74 & .73 & .76 & .63 & .57 & .63\\ \hline
        IsoMap & .65 & .78 & .77 & .79 & .54 & .48 & .55\\ \hline
    \end{tabular}
    \caption{Comparison of the different manifold learning embeddings considered with different classifiers.}
    \label{tab:manifold_learning_comparison}
    \vspace{-0.2 cm}
\end{table}

\subsection{Semi-Supervised Learning}

\subsubsection{Co-Training}
For co-training, we first split the dataset into labeled and unlabeled sets randomly in the ratio of 1:2. In the real world, this randomness may be caused due to accidental cessation of the Snort application or if a network security expert cannot make an inference of intrusion. Further both the labeled and unlabeled data are split into cyber and physical views consisting of respective features. In these experiments, we compare the supervised learning techniques on the labeled dataset with the co-training technique which uses supervised learning cyber and physical classifiers as shown in Fig.~\ref{fig:cotraining}. It is expected to have a reduction in performance from supervised learning techniques, due to lack of labels for some samples, but it can be observed from Table~\ref{tab:cotrain_comparison}, that the co-training based classification outperforms supervised for some classifiers such as $LR$,$GNB$,$BNB$,$MLP$ and performs at par with other classifiers with a difference of a mere 8 percent in the case of $RF$. The probable reason for improvement in performance using co-training may be due to the training of two different classifiers using intra-domain features.

\begin{table}[h]
    \centering
    \begin{tabular}{|l|l|l|l|l|l|l|}
    \hline
    Classifier & \multicolumn{3}{|c|}{Supervised} & \multicolumn{3}{|c|}{Co-Training}\\ \hline
          & F1-score & Rec.& Prec.& F1-score & Rec.& Prec.\\ \hline
        LR & .63 & .67 & .64 & .64 & .73 & .58\\ \hline
        SVC & .63 & .67 & .64 & .59 & .70 & .52\\ \hline
        DT & .69 & .71 & .69 & .64 & .71 & .65 \\ \hline
        RF & .73 & .77 & .72 & .65 & .72 & .72\\ \hline
         GNB & .28 & .33 & .66 & .30 & .32 & .56\\ \hline
        BNB & .53 & .51 & .67 & .58 & .66 & .52\\ \hline
        MLP  & .59 & .71 & .51 & .61 & .71 & .55\\ \hline
    \end{tabular}
    \caption{Comparison of the classifier using supervised and co-training based unsupervised learning.}
    \label{tab:cotrain_comparison}
    \vspace{-0.2 cm}
\end{table}

%% file: Content/10_conclusion.tex
\section{Conclusion}\label{conclusion}
A data fusion framework for detecting false command and measurement injections due to cyber intrusion is presented in this paper. To design an IDS that uses cyber and physical features, we aggregate features from cyber and physical sensors and align the data, then perform preprocessing techniques, followed by inter-domain fusion. 

Our results find that classifier performance improves on an average of 15- 20\% (based on F1-score) when cyber physical features are considered instead of pure cyber features. Results also show that the performance improved on an average of 10-20\% (based on F1-score) when labels from Snort are replaced by the labels considered based on intrusion timestamps.  From our evaluations of the IDS, we also find that scenarios with balanced and larger records result in better performance.  Additionally,  co-training based semi-supervised learning technique, which is realistic for a real-world scenario, is found to perform similar to supervised techniques and even better by 2-5\% (based on F1-score) using some classifiers. Among the unsupervised learning techniques, k-mean clustering technique is found to be more robust and accurate. Moreover, training the classifier with the embeddings from manifold learning didn't improve the accuracy. Hence, manifold learning should only be considered for visualization rather than rely on accuracy.

We believe our fused dataset and results provide one of the first publicly available studies with cyber and physical features, particularly for power systems, where the experimental data is collected from a testbed that contains both cyber and physical emulation. This benefits research in multi-disciplinary areas such as cyber physical security and data science.